\documentclass[]{elsarticle}

\usepackage{hyperref}
\usepackage{subfig}
\usepackage{graphicx}
\usepackage{algorithm}% http://ctan.org/pkg/algorithms
\usepackage{algpseudocode}% http://ctan.org/pkg/algorithmicx
\usepackage[]{amsmath}
\usepackage{url}
\usepackage[section]{placeins} % figure cannot float to next section
\newcommand\norm[1]{\left\lVert#1\right\rVert}
\usepackage[colorinlistoftodos]{todonotes}

\journal{Robotics and Autonomous System}

\begin{document}

\begin{frontmatter}

\title{Autonomous drone race:
A computationally efficient vision-based navigation and control strategy}


\author[add1,corr]{S.Li}
\cortext[corr]{Corresponding author}
\ead{s.li-4@tudelft.nl}
\author[add1]{M.M.O.I. Ozo}
\author[add1]{C. De Wagter}

\author[add1]{G.C.H.E. de Croon}

\address[add1]{Micro Air Vehicle lab, Control and Simulation, Delft University of Technology, The Netherlands}

\begin{abstract}
Drone racing is becoming a popular sport where human pilots have to control their drones to fly at high speed through complex environments and pass a number of gates in a pre-defined sequence. 
In this paper, we develop an autonomous system for drones to race fully autonomously using only onboard resources.
Instead of commonly used visual navigation methods, such as simultaneous localization and mapping and visual inertial odometry, which are computationally expensive for micro aerial vehicles (MAVs), we developed the highly efficient snake gate detection algorithm for visual navigation, which can detect the gate at 20HZ on a Parrot Bebop drone.
Then, with the gate detection result, we developed a robust pose estimation algorithm which has better tolerance to detection noise than a state-of-the-art perspective-n-point method. During the race, sometimes the gates are not in the the drone's field of view. For this case, a state prediction-based feed-forward control strategy is developed to steer the drone to fly to the next gate. Experiments show that the drone can fly a half-circle with 1.5m radius within 2 seconds with only $30cm$ error at the end of the circle without any position feedback. Finally, the whole system is tested in a complex environment (a showroom in the faculty of Aerospace Engineering, TU Delft). The result shows that the drone can complete the track of 15 gates with a speed of $1.5m/s$ which is faster than the speeds exhibited at the 2016 and 2017 IROS autonomous drone races.
\end{abstract}


\end{frontmatter}

%\linenumbers

\section{Introduction}

First person view (FPV) drone racing has been a popular sport in recent years, where the pilots have to control the drones to fly through gates decorated by LED lights at fast speed. In the field of robotics, drone racing has raised the question: how can drones be designed to fly races by themselves, possibly faster than human pilots? To start answering this question, the world's first autonomous drone race was held in 2016 \cite{Moon2017}. In this drone race, the drones were asked to fly through square, orange gates in a predefined sequence using onboard resources. To increase the level of challenge for gate detection, unlit gates were used in the race. The rules were simple: the one who flies furthest will win the race, and if two drones arrive at the same gate or complete the full track, the fastest time counts. The winner of the 2016 race (the team from KAIST) flew through 10 gates (the distance is around $50m$) within 86s \cite{jung2018direct} and the winner of the 2017 race (the team from INAOE) flew through 9 gates ($60m$) within 194s, which are much slower than the FPV drone race players. Compared to the FPV drone race, the task of autonomous drone race is more challenging because the drone has to navigate, perceive, plan and control all by itself using only scarce onboard resources, representing a considerable challenge for areas such as artificial intelligence and control.

Autonomous drone racing can be seen in the more general context of high-speed flight. In fact, before the autonomous drone race, there are several on flying through circles or gaps. To the best of our knowledge, the first research on quadrotor's flying through circles is \cite{Mellinger2011}. In their work, the drone can fly through a thrown circle and three fixed circles with fast speed. In \cite{Mellinger2012}, the drone can fly through a tilted narrow gap. In both studies, a VICON motion capture system is used to provide the state estimation for the drone and the position of the gap or circles is known a priori. Lyu et al. \cite{Lyu2015} use an onboard camera to detect the gap and the drone could navigate itself through the gate. But the image processing is done off-board. In their experiment, the background of the gap is a white wall which makes the gap to be detected relatively easily. Loianno et al. \cite{loianno2017estimation} for the first time use onboard resources to detect a window, plan the trajectory and control the drone to fly through a window. In their work, visual inertial odometry (VIO), which is computationally quite expensive for our drone, is used to provide the state estimation to the drone. In Falanga et al.'s \cite{Falanga2017} work, a drone with a fish-eye camera can detect a black and white gap and design a trajectory through the gap using only onboard resources. In \cite{sanket2018gapflyt}, deep-learning-based optical flow is used to find any arbitrary shaped gap with an NVIDIA Jetson TX2 GPU. But the drone has to execute a fixed sideways translational motion to detect the gap before going through it, which slows down the drone. The studies above aim at motion planning, object detection or onboard perception, so in most of these studies only one gap is flown through and there is no solution on how to fly through the next gate after passing through the previous one. 

Multiple studies have focused directly on autonomous drone racing, designing a strategy that will allow to fly an entire trajectory. In \cite{li2018teaching}, a simulated drone learns how to minimize the time spent to finish the race track, by learning from two different PID controllers. Although an interesting approach, it ignores several of the real-world aspects of drone racing, such as restricted onboard computation or how to deal with accelerometer biases. NASA's Jet Propulsion Laboratory has developed an autonomous racing drone controlled by AI, which can fly almost as fast as the racing drones controlled by expert human FPV pilots.\cite{good2017drone,morrell2018differential} They use VIO for navigation which is computationally relatively expensive. Kaufmann et al. develop a strategy that combines a convolutional neural network (CNN) and minimum jerk trajectory generation.\cite{kaufmann2018deep} In their work, an in-house quadrotor with an Intel UpBoard and a Qualcomm Snapdragon Flight Kit which is used for VIO, is used as the platform. In \cite{jung2018direct}, a systematic solution for the IROS autonomous drone race 2016 is presented. In their work, an NVIDIA Jetson TK1 single-board computer and a stereo camera are used for a visual servoing task. They finally passed through 10 gates within 86s and won the race. We will use their result as a benchmark to compare our research result. 

In this paper, we present a solution for autonomous drone racing, which is computationally more efficient than the solutions discussed above. For the gate detection, a novel light-weight algorithm, ``snake gate detection'', is described and analyzed in detail in Section~\ref{sec:vision_navigation}. Instead of using a common, purely vision-based perspective-n-point (PnP) algorithm, we combine the onboard attitude estimate with the gate detection result to determine the position of the drone. We show that this is more robust than the PnP method. Then, a novel Kalman filter is introduced that uses a straightforward drag model to estimate the velocity of the drone. Two control strategies to control the drone to go through the gate and find the next gate are discussed in Section~\ref{sec:control_strategy}. In Section \ref{sec:experiment}, flight tests are performed with a Parrot Bebop~1 drone, by replacing the Parrot firmware with our Paparazzi autopilot code. All algorithms run in real-time on the limited Parrot P7 dual-core CPU Cortex A9 processor, and no hardware changes are required as the vision algorithms use the frontal camera and other sensors already present in the Bebop. The flight experiments are done in a complex and narrow environment (a showroom displaying aircraft components in the basement of Aerospace Engineering, TU Delft).\footnote{The video of the experiment is available at: https://youtu.be/bwF0TAjC8iI} The result shows that the drone can fly through a sequence of 15 gates autonomously using only onboard resources in a very complex environment with a velocity of up to $1.5m/s$.

\section{System overview}

\begin{figure}[h]
    \centering
    \includegraphics[width=0.75\textwidth]{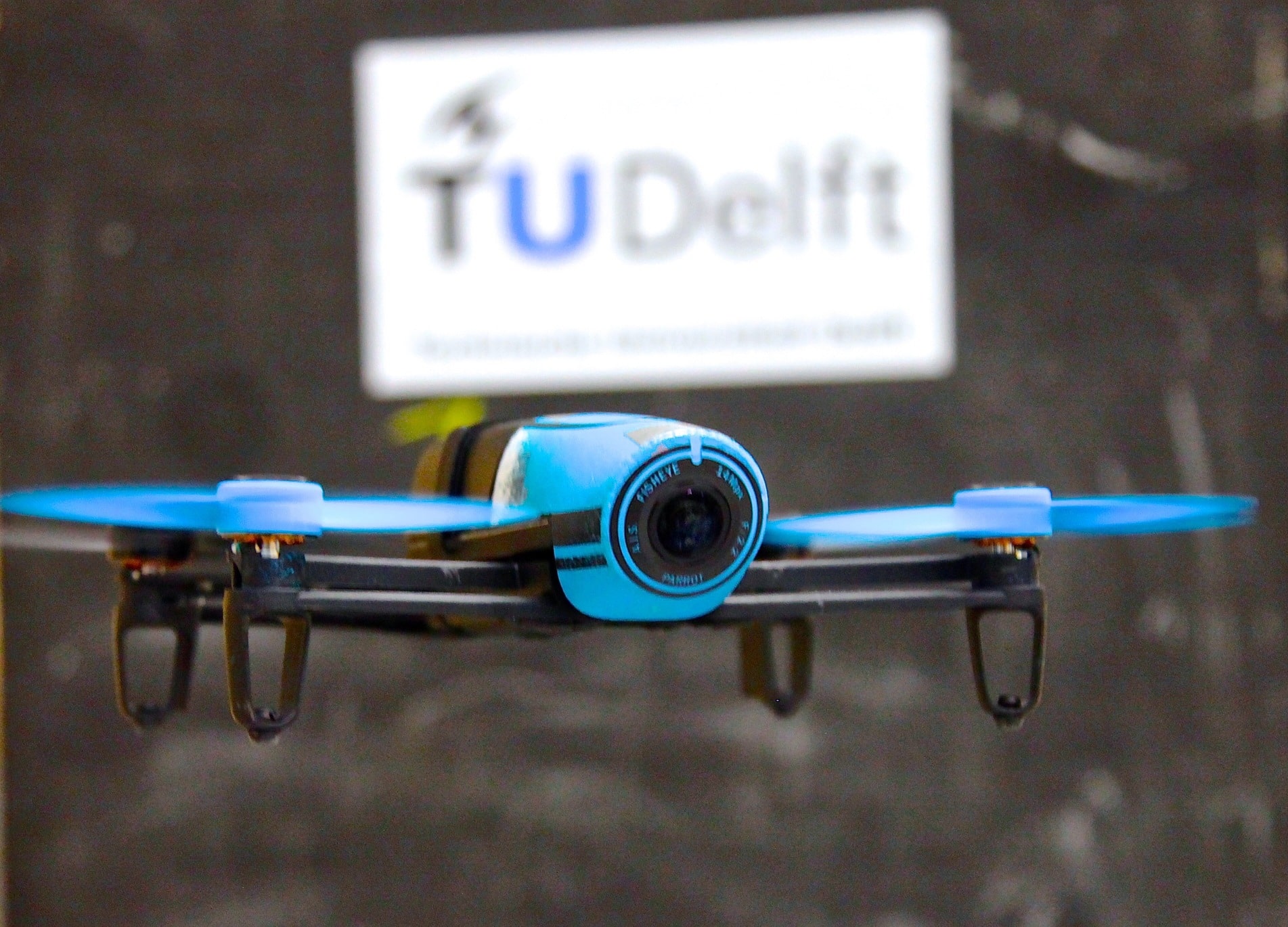}
    \caption{The Parrot Bebop 1 is used as experiment platform. The software is replaced by the Paparazzi UAV open-source autopilot project}
    \label{fig:bebop_cyberzoo}
\end{figure}

The quadrotor hardware used as experiment platform in this work is a commercially available Parrot Bebop~1 (Figure \ref{fig:bebop_cyberzoo}). However, all Parrot software was replaced by own computer vision, own sensor drivers and own navigation and control using the Paparazzi-UAV open-source autopilot project.\cite{gati2013open} Only the Linux operating system was kept.  The most important characteristics are listed in Table \ref{tab:list of sensors}. It should be noted that the image from the front camera as used by our autopilot in this work is only $160\times350$ pixels and all the processing for the drone race takes place on the Parrot P7 dual-core CPU Cortex 9 (max 2GHz), although the Bebop is equipped with a quad core GPU.

\begin{table}[h]
\caption{List of onboard sensors used in the experiment} % title of Table
\centering % used for centering table
\begin{tabular}{c c} % centered columns (4 columns)
\hline\hline %inserts double horizontal lines
camera & a 6 optical elements and 14 Mega pixels sensor \\
{} &  a vertical stabilization camera (not used in this work)  \\
 processor & Parrot P7 dual-core CPU cortex 9 (max 2GHz) \\  
 IMU & MPU 6050  \\
 sonar & $<8m$  \\[1ex] % [1ex] adds vertical space
 \hline
\end{tabular}
\label{tab:list of sensors} % is used to refer this table in the text
\end{table}

\begin{figure}[h]
    \centering
    \includegraphics[width=\textwidth]{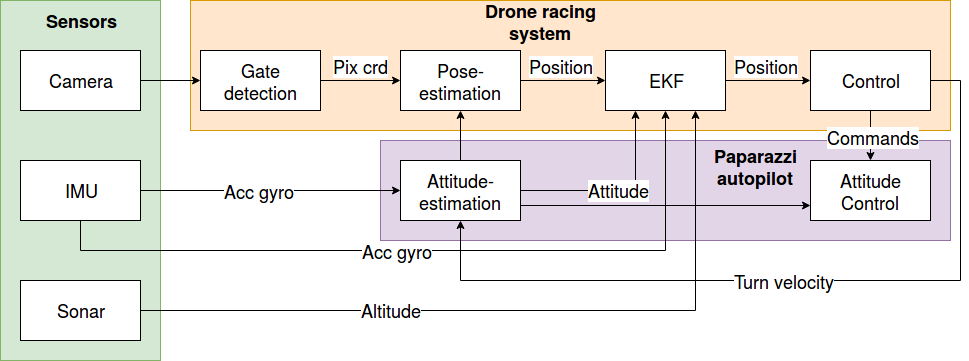}
    \caption{The structure of the autonomous system}
    \label{fig:system_structure}
\end{figure}

The structure of the system is shown in Figure \ref{fig:system_structure}. For visual navigation, a novel algorithm, snake gate detection, is implemented to detect the gates. It outputs the coordinates of detected gates' corners, which are then sent to the pose estimation block. In pose estimation block, the coordinates of the gate corners on the image plane would be projected to 3D space, which provides the relevant position between the drone and the gate. For attitude and heading reference system (AHRS), a classic complementary filter \cite{Euston2008} is employed. At last, the position measured by the front camera, attitude estimation from AHRS and IMU measurement are fused by a Kalman filter to provide a position estimate. 

In terms of control, when the target gate is in the field of view, a PD controller (Control block in Figure \ref{fig:system_structure}) is used to steer the drone to align with the center of the gate. After passing through the gate or there is no gate in the field of view, a prediction-based feed-forward control scheme is employed to steer the drone to the next gate, which will be further explained in Section 4. An adaptive incremental nonlinear dynamic inversion (INDI) controller is used as low-level attitude controller \cite{Smeur2016}. 

\begin{figure}[h]
    \centering
    \includegraphics[trim= 40 10 40 10,clip,scale =0.5]{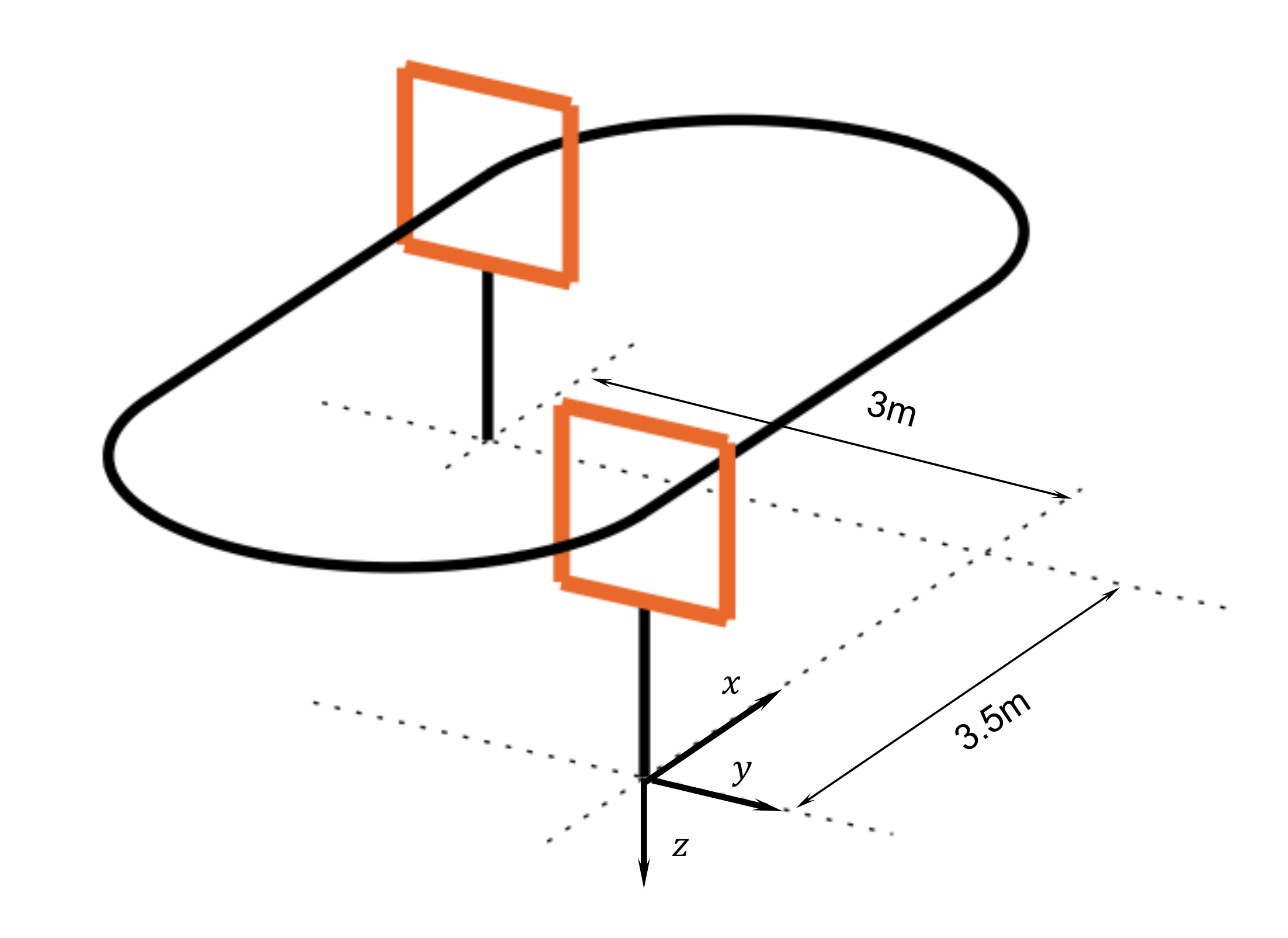}
    \caption{A simplified race track}
    \label{fig:race track}
\end{figure}

The race track can be divided into two parts. The first part is the approaching gate part where the target gate can be used by the drone for navigation. The other one is after gate part, which starts from the point where the drone passing through the gate and ends at the point where the drone can see the next gate. The different race tracks can be seen as the different combination of these two parts. Thus, at first , due to the space restriction of our experimental environment, we simplify the race track to a two gates track which can be seen in Figure \ref{fig:race track}. Most of our experiments are done and analyzed in this simplified race track with the ground truth measurement provided by Opti-track. At last, the system is moved to a more complex and realistic drone race track to be verified.  

\section{Vision navigation}\label{sec:vision_navigation}
In the FPV drone race, gates are usually decorated with LEDs in order to be easily recognized by drone pilots. Drone pilots can then use the gates to navigate themselves to approach the gates. Inspired by FPV drone race, in our research, we also use gates for navigation since their simple shape and relatively large size make them relatively easy to be extracted and their projection on the image plane can provide information such as position and attitude of the drone. In this section, we first present an efficient gate detection method to extract the four corners of the gate on the image plane. Next, the position of the four corners of the gate is projected to 3D space combining AHRS reading. At last, a Kalman filter providing position estimation by fusing the vision measurement, the IMU measurement and the onboard AHRS reading is discussed.

\subsection{Gate detection}
Gate detection can be accomplished by multiple different computer vision methods, such as Viola and Jones\cite{Viola2001}, Hough transform\cite{Illingworth1988} and deep learning\cite{Ren2015,Jung20188}. In this article, we propose a novel gate detection algorithm called snake gate detection which is lightweight and easy to be implemented onboard.

\begin{figure}[!h]
\centering
\subfloat[If the gate is continuous on image plane, snake gate detection algorithm should find all four corners $P_1$, $P_2$, $P_3$ and $P_4$]
{
  \includegraphics[width=0.4\textwidth]{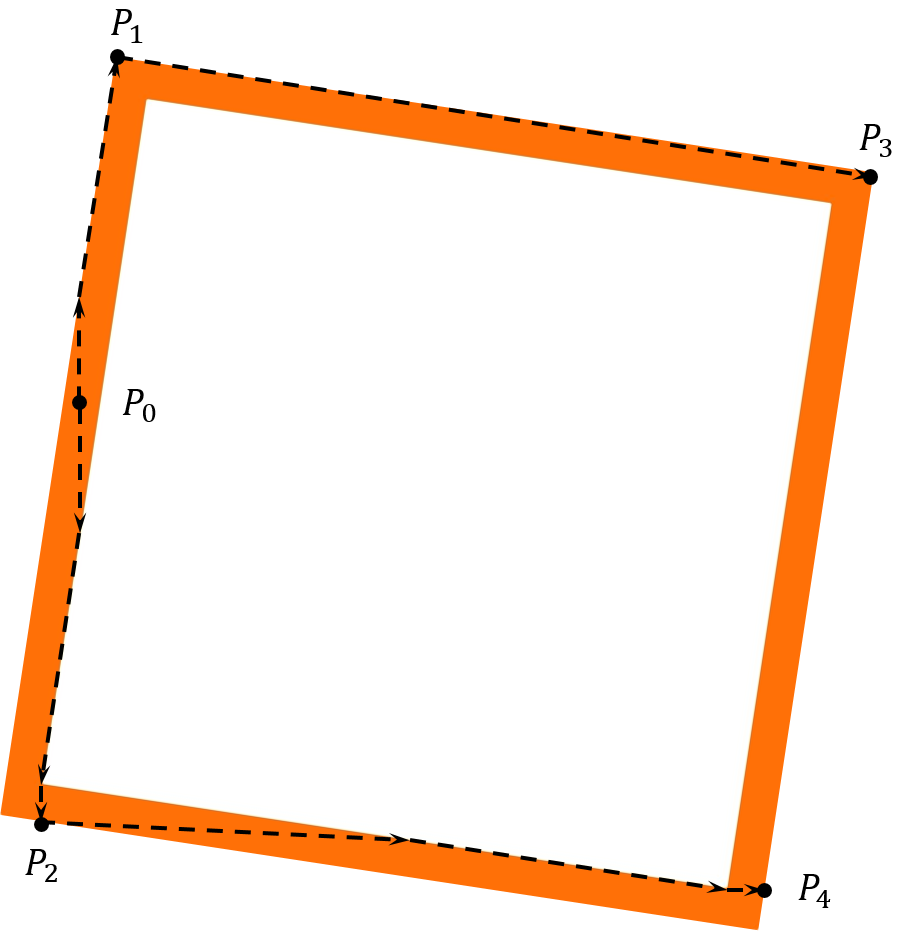}
  \label{fig:raw snake gate}
}
 \hfill
\subfloat[ When the gate is not continuous on image plane, first a square $S_1$, $S_2$, $S_3$ , $S_4$ with minimum length including $P_1$, $P_2$, $P_3$ , $P_4$ is found. Four small squares centering at $S_i$ are then found. In these small squares, a histogram analysis helps to refine our estimate of the gate's corners in the image]
{
  \includegraphics[width=0.4\textwidth]{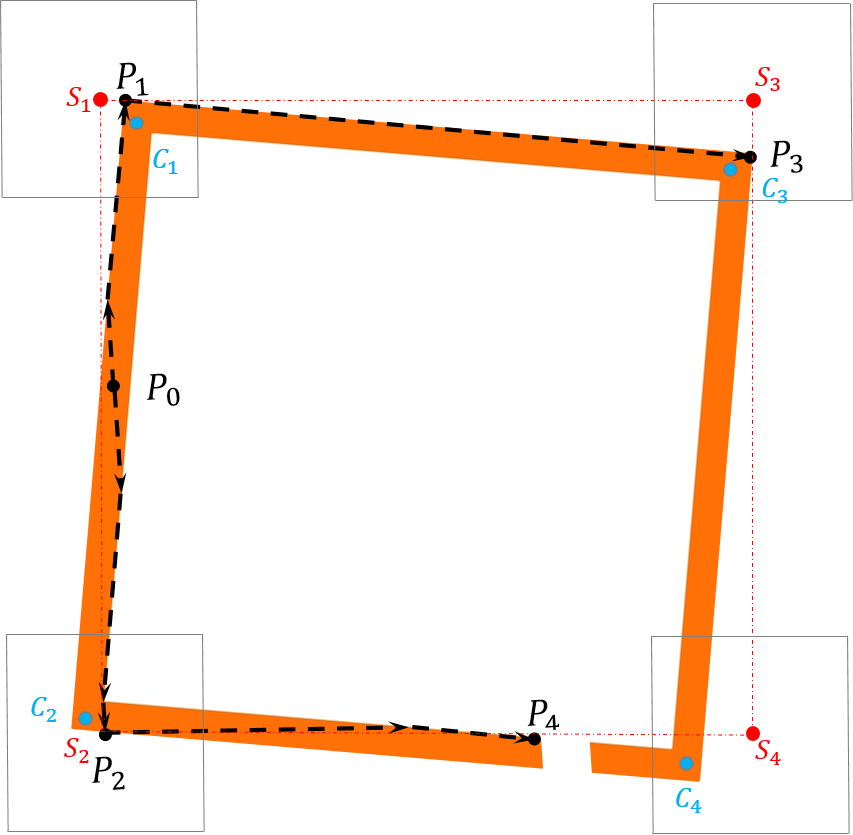} \label{fig:snake overexposure}
}
\caption{An example of snake gate detection.}
\label{fig:snake gate detection demo}
\end{figure}

We search the gates based on their colors on an distorted image because the undistortion procedure for each image can slow down the whole detection procedure. (Figure \ref{fig:snake gate detection demo}) Luckily, our detection method can still work properly on this distorted image. The search starts by randomly sampling \cite{de2012sub} in the original image. If a random point $\mathbf{P}_0$ hits the target color (gate's color), we continue searching 'up and down' to find points $\mathbf{P}_1$ and $\mathbf{P}_2$. It should be noted that this search can search along the edge of the oblique bar of the gate. (Figure \ref{fig:raw snake gate}) To prevent that the algorithm may find some small color blocks which have the same color as the gate, we introduce a threshold, which is called the minimum length threshold $\sigma_L$. If $\lVert\mathbf{P}_1-\mathbf{P}_2\rVert < \sigma_L$, this search would be terminated. Then, we use $\mathbf{P}_1$ and $\mathbf{P}_2$ as start points respectively to search 'left and right' to find $\mathbf{P}_3$ and $\mathbf{P}_4$. Similar to the vertical search, the horizontal search can also search along the oblique bar and the result would be checked by $\sigma_L$ to ensure that the detection is not too small and hence unlikely to be a gate. The algorithm can be found in Algorithm \ref{alg:snake gate detection}. It should be noted that while small $\sigma_L$ may lead to acceptance of some small detections which in most cases are false positive detections, large $\sigma_L$ can lead to the result that some gate in the image  are rejected. The selection of $\sigma_L$ will be discussed later in this section. 

If the gate's image is continuous in the image plane and the gates' edges are smooth, snake gate detection should find all four points. (Figure \ref{fig:raw snake gate}) However, due to varying light conditions, some parts of the gate may get overexposure or underexposed which may lead to color deviation. For example, in Figure \ref{fig:snake overexposure}, part of the lower bar gets overexposed. In this case, $\mathbf{P}_4$ will not reach the real gate's corner. Hence, a refining process is employed to find the real gate's corner. To refine the detection, a square with minimum length including four points is firstly obtained. (Red square in Figure \ref{fig:snake overexposure}) Then four small squares centering at $S_i$ are found.(Four gray square in Figure \ref{fig:snake overexposure}) The raw detection is refined by finding the centroid of the patch around each rough corner. 

In one image, in most cases, the number of detected gates $N_d$ is larger than the number of real gates in the image $N_g$. It can be caused by duplicated samples on the same gate, which are true positive detections and do not affect the performance of navigation. The other reason for $N_d > N_g$ is the false positive detections, which affects the accuracy of navigation significantly and should be eliminated. Here, another threshold, color fitness threshold $\sigma_{cf}$, is introduced to help decrease the number of false positive detection. 

\begin{align}
cf = \frac{N_c}{N}
\end{align}

where $N_c$ is the total number of pixels on the polygon whose color is target color and $N$ is the number of pixels on the polygon.

Only the gates whose $cf > \sigma_{cf}$ will be accepted as detected gates. Similar to minimum length threshold $\sigma_L$, the selection of $\sigma_{cf}$ also affects the detection accuracy significantly. 

\begin{algorithm}[h!]
  \caption{snake gate detection}
  \begin{algorithmic}[1]
    \Procedure{SnakeGateDetection}{$image$}  %\Comment{The g.c.d. of a and b}
      \For{\texttt{i = 1:maxSample}}
         \State $\mathbf{P}_0$ = randomPoint()
         \If{isTargetColor($\mathbf{P}_0$,$image$)}
                \State [$\mathbf{P}_1$,$\mathbf{P}_2$] = searchUpDown($\mathbf{P}_0$,$image$)
                \If{$\lVert\mathbf{P}_1-\mathbf{P}_2\rVert > \sigma_L$}
                \State $\mathbf{P}_3$ = searchLeftRight($\mathbf{P}_1$,$image$)
                \State $\mathbf{P}_4$ = searchLeftRight($\mathbf{P}_2$,$image$)
                \If{$\lVert\mathbf{P}_1-\mathbf{P}_3\rVert >\sigma_L$  OR $\lVert\mathbf{P}_2-\mathbf{P}_4\rVert >\sigma_L$}
                \State  [$\mathbf{S}_1$,$\mathbf{S}_2$,$\mathbf{S}_3$,$\mathbf{S}_4$] = findMinimalSquare($\mathbf{P}_1$,$\mathbf{P}_2$,$\mathbf{P}_3$,$\mathbf{P}_4$)
                \State detectedGate = refineCorner($\mathbf{S}_1$,$\mathbf{S}_2$,$\mathbf{S}_3$,$\mathbf{S}_4$)
                \If{checkColorFitness(detectedGate) $> \sigma_{cf}$}
                \State \textbf{return} $detectedGate$
                \EndIf
                \EndIf
                \EndIf
          \EndIf
      \EndFor
      
    \EndProcedure
  \end{algorithmic}
  \label{alg:snake gate detection}
\end{algorithm}

\begin{algorithm}[h!]
  \caption{search in vertical direction (search in horizontal direction is similar)}\label{euclid}
  \begin{algorithmic}[1]
    \Procedure{searchUpandDown}{$\mathbf{P0},image$}  %\Comment{The g.c.d. of a and b}     
         \State $\mathbf{P}_1$ = $\mathbf{P}_0$, $\mathbf{P}_2$ = $\mathbf{P}_0$, $done$ = $false$
         \While{!$done$} 
         \If{isTargetColor($\mathbf{P}_1.x,\mathbf{P}_1.y-1)$} 
         \State $\mathbf{P}_1.y = \mathbf{P}_1.y-1$
         \ElsIf{isTargetColor($\mathbf{P}_1.x-1,\mathbf{P}_1.y-1)$} 
         \State  $\mathbf{P}_1.x = \mathbf{P}_1.x-1$
         \State  $\mathbf{P}_1.y = \mathbf{P}_1.y-1$
         \ElsIf{isTargetColor($\mathbf{P}_1.x+1,\mathbf{P}_1.y-1)$} 
         \State  $\mathbf{P}_1.x = \mathbf{P}_1.x+1$
         \State  $\mathbf{P}_1.y = \mathbf{P}_1.y-1$
         \Else
         \State $done = true$
         \EndIf
         \EndWhile 
         \State $done = false$
          \While{!$done$} 
         \If{isTargetColor($\mathbf{P}_2.x,\mathbf{P}_1.y+1)$} 
         \State $\mathbf{P}_2.y = \mathbf{P}_2.y+1$
         \ElsIf{isTargetColor($\mathbf{P}_2.x-1,\mathbf{P}_1.y+1)$} 
         \State  $\mathbf{P}_2.x = \mathbf{P}_2.x-1$
         \State  $\mathbf{P}_2.y = \mathbf{P}_2.y+1$
         \ElsIf{isTargetColor($\mathbf{P}_1.x+1,\mathbf{P}_1.y+1)$} 
         \State  $\mathbf{P}_2.x = \mathbf{P}_2.x+1$
         \State  $\mathbf{P}_2.y = \mathbf{P}_2.y+1$
         \Else
         \State $done = true$
         \EndIf
         \EndWhile 

      \State \textbf{return} $\mathbf{P}_1$, $\mathbf{P}_2$   %\Comment{The gcd is b}
    \EndProcedure
  \end{algorithmic}
\end{algorithm}

\begin{figure}

\centering
\subfloat[Ture positive detection]{\includegraphics[trim= 40 10 40 10,clip,width=4cm,height=3cm]{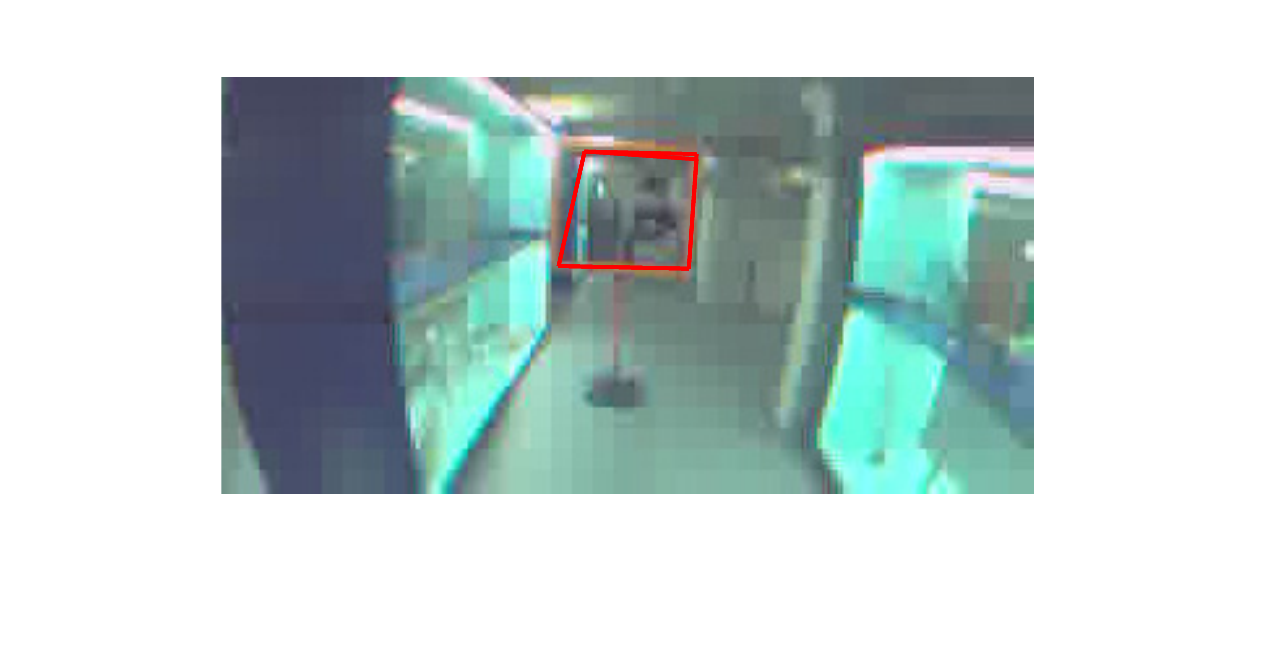}}\hfil
\subfloat[Ture positive detection and false positive detection]{\includegraphics[trim= 40 10 40 10,clip,width=4cm,height=3cm]{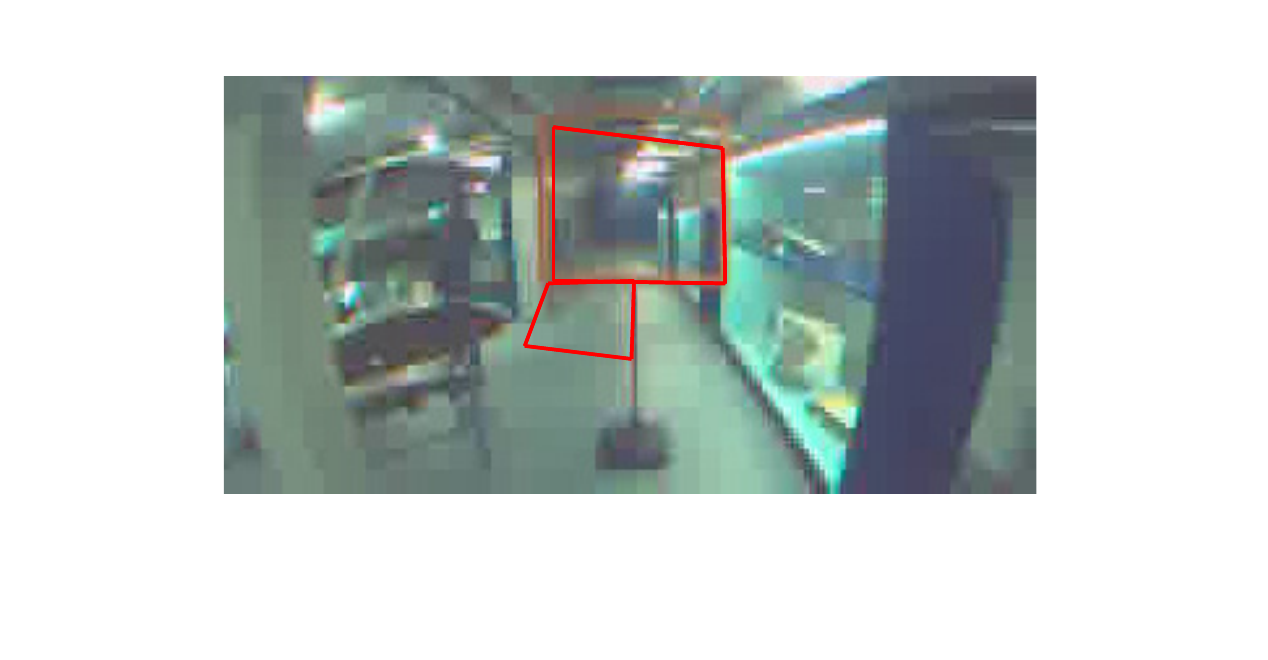}}\hfil 
\subfloat[False negative detection]{\includegraphics[trim= 40 10 40 10,clip,width=4cm,height=3cm]{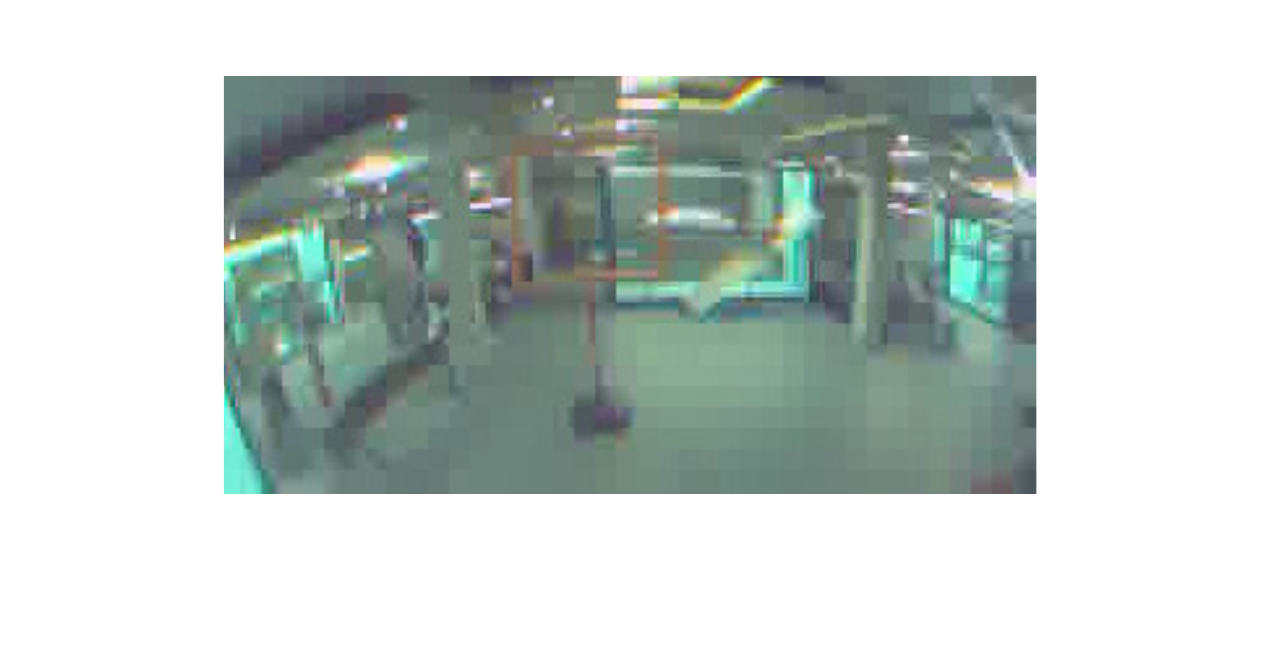}} 

\subfloat[]{\includegraphics[width=4cm]{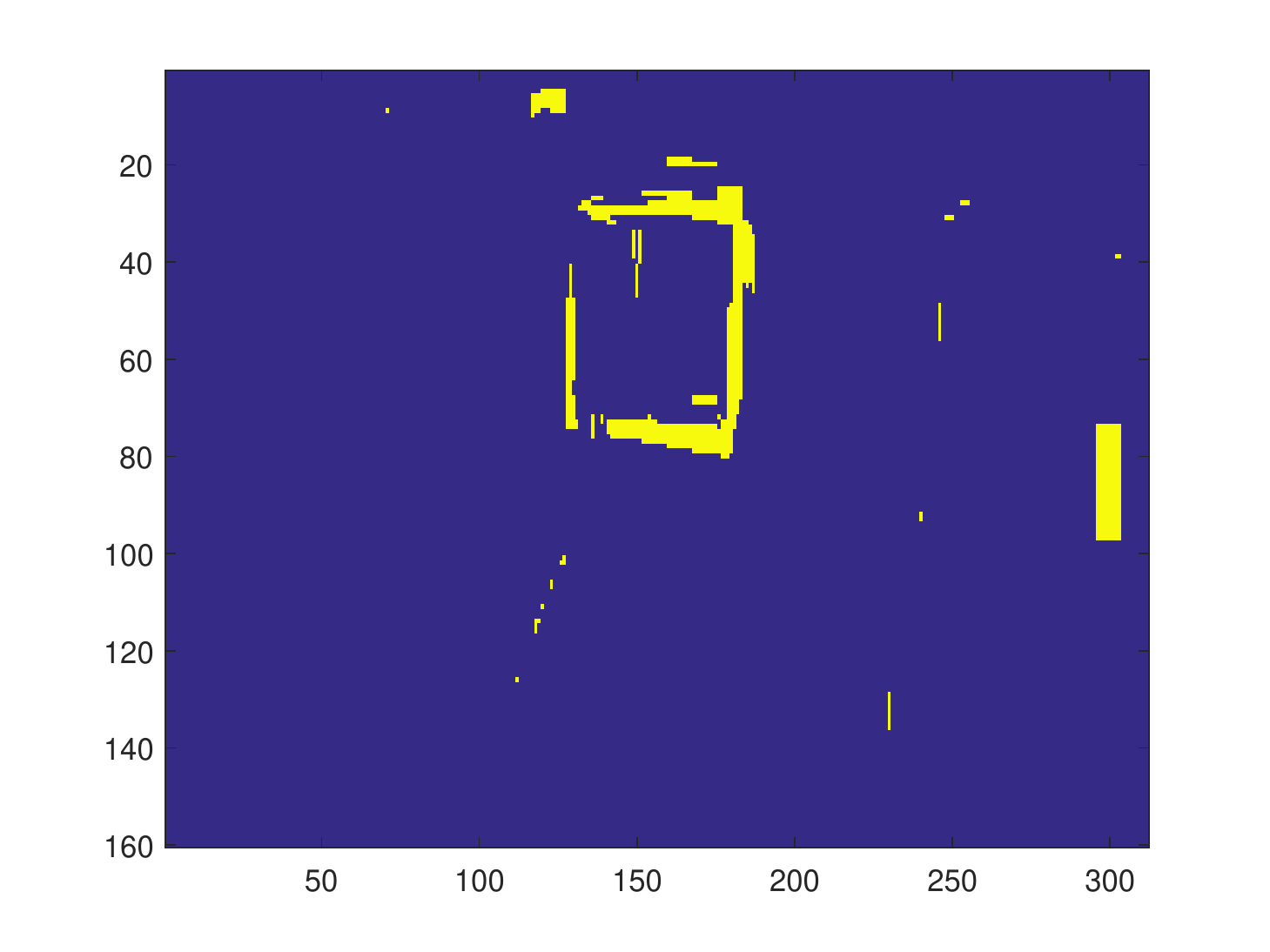}}\hfil   
\subfloat[]{\includegraphics[width=4cm]{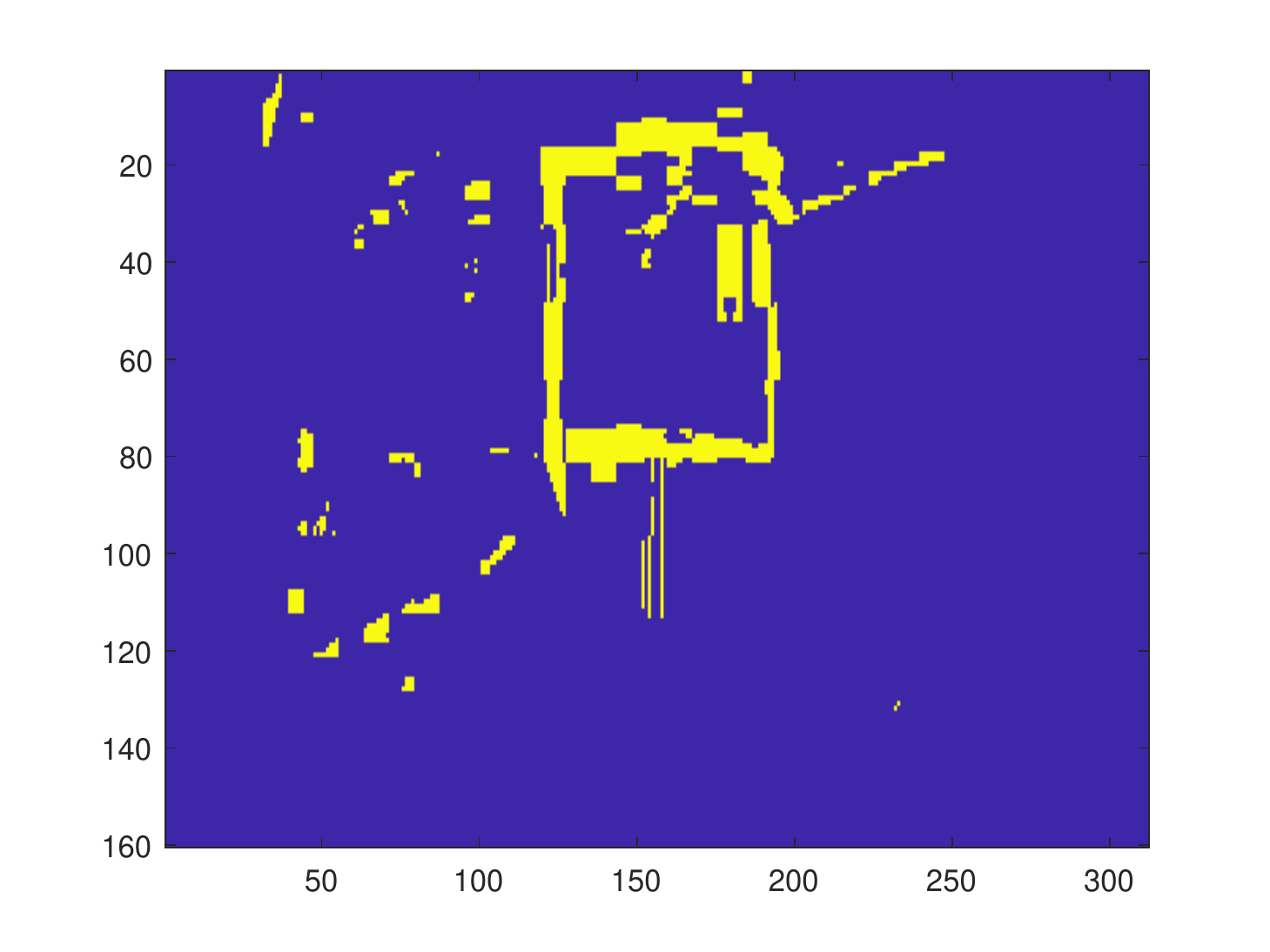}}\hfil
\subfloat[]{\includegraphics[width=4cm]{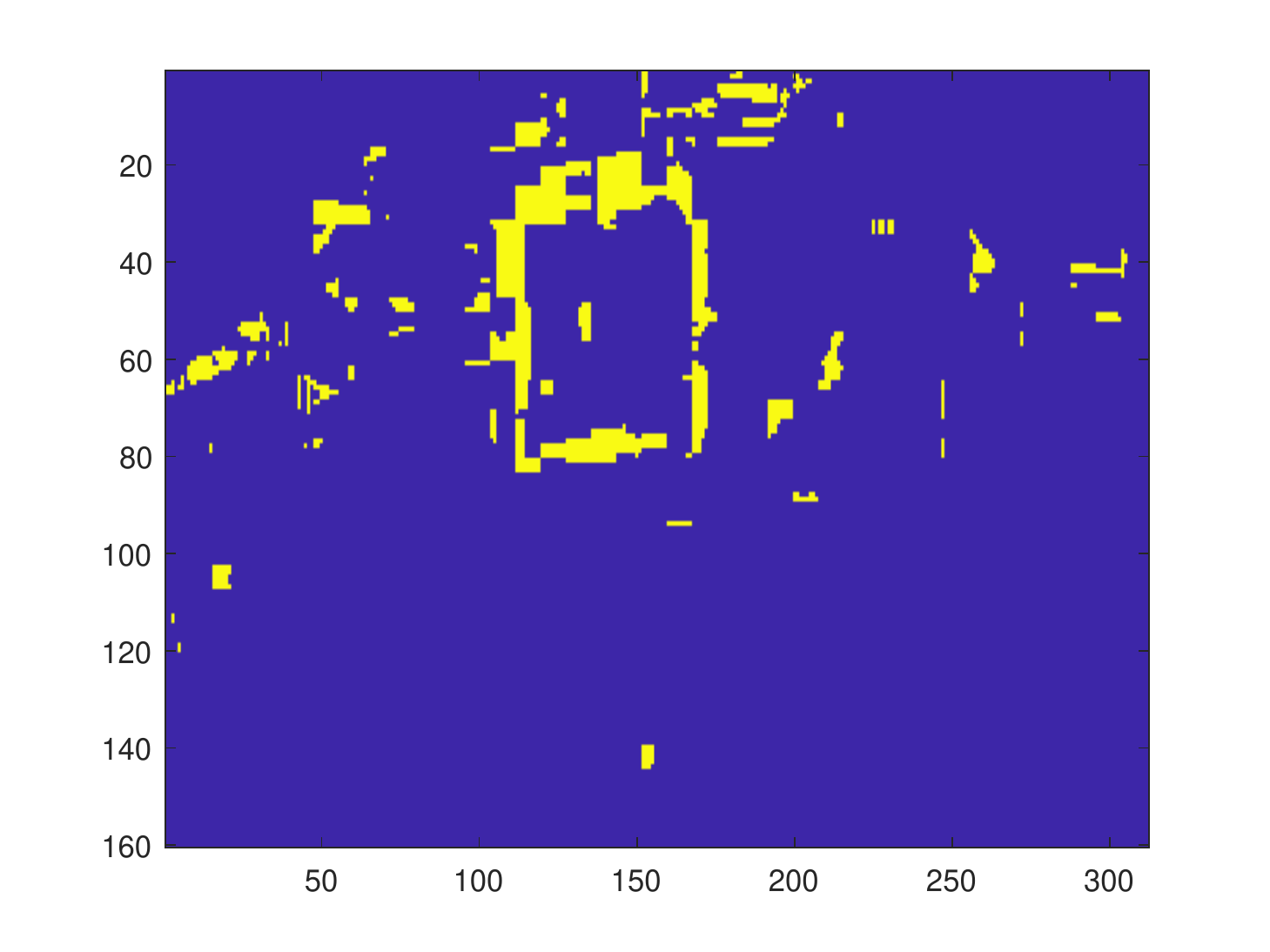}}
\caption{Examples of the snake gate detection results. The first row are original onboard images with detection results. The second row are corresponding masks}\label{fig:ROC examples}
\end{figure}

\begin{figure}[!h]
\centering
\subfloat[]
{
  \includegraphics[scale=.4]{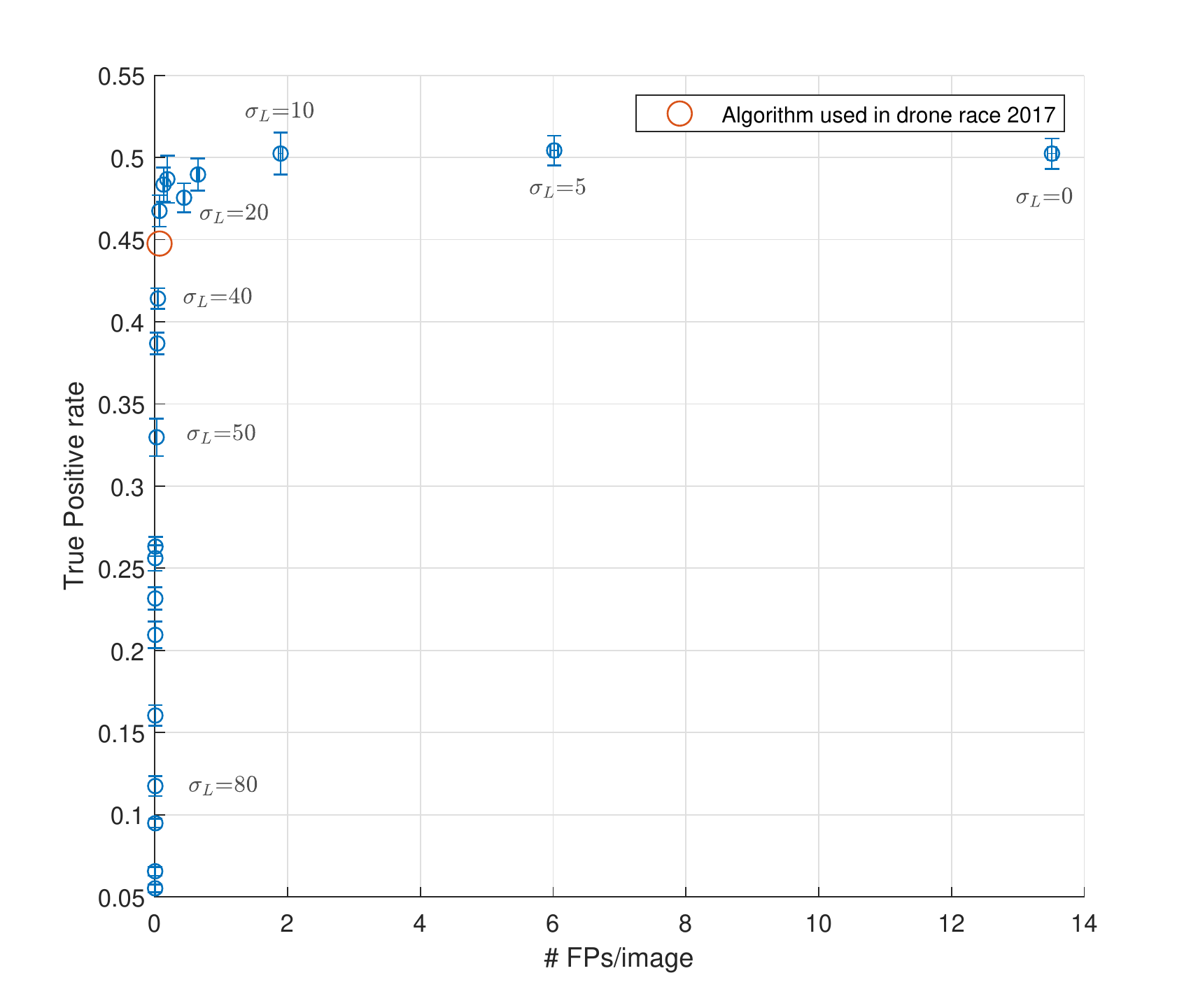} \label{fig:original ROC}
}
\hfill
\subfloat[]
{
  \includegraphics[scale=.4]{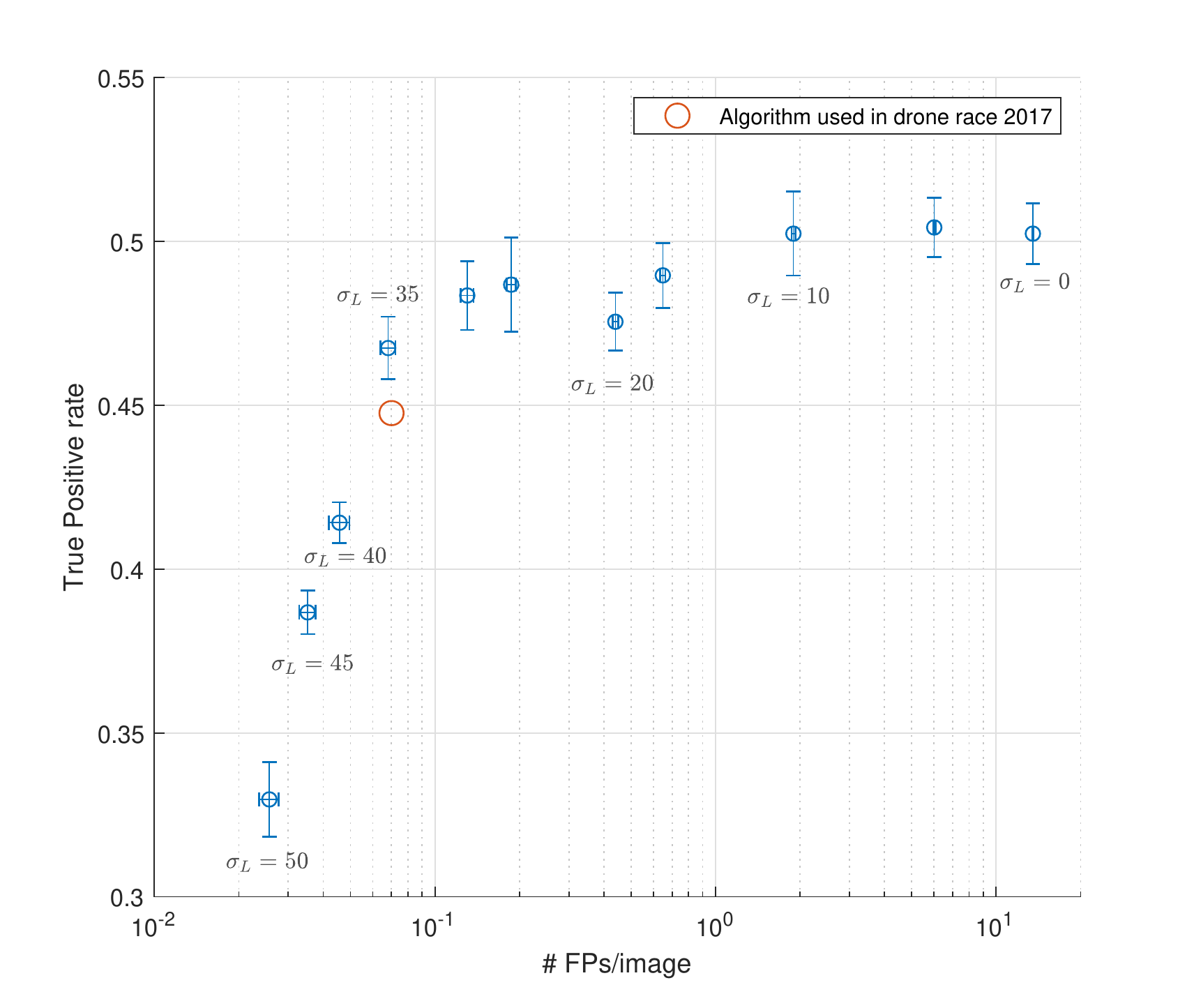} \label{fig:overexposure} \label{fig:local ROC}
}
\caption{The ROC curve with the change of $\sigma_L$.}
\label{fig:ROC min length}
\end{figure}

\begin{figure}[h]
    \centering
    \includegraphics[width=0.6\textwidth]{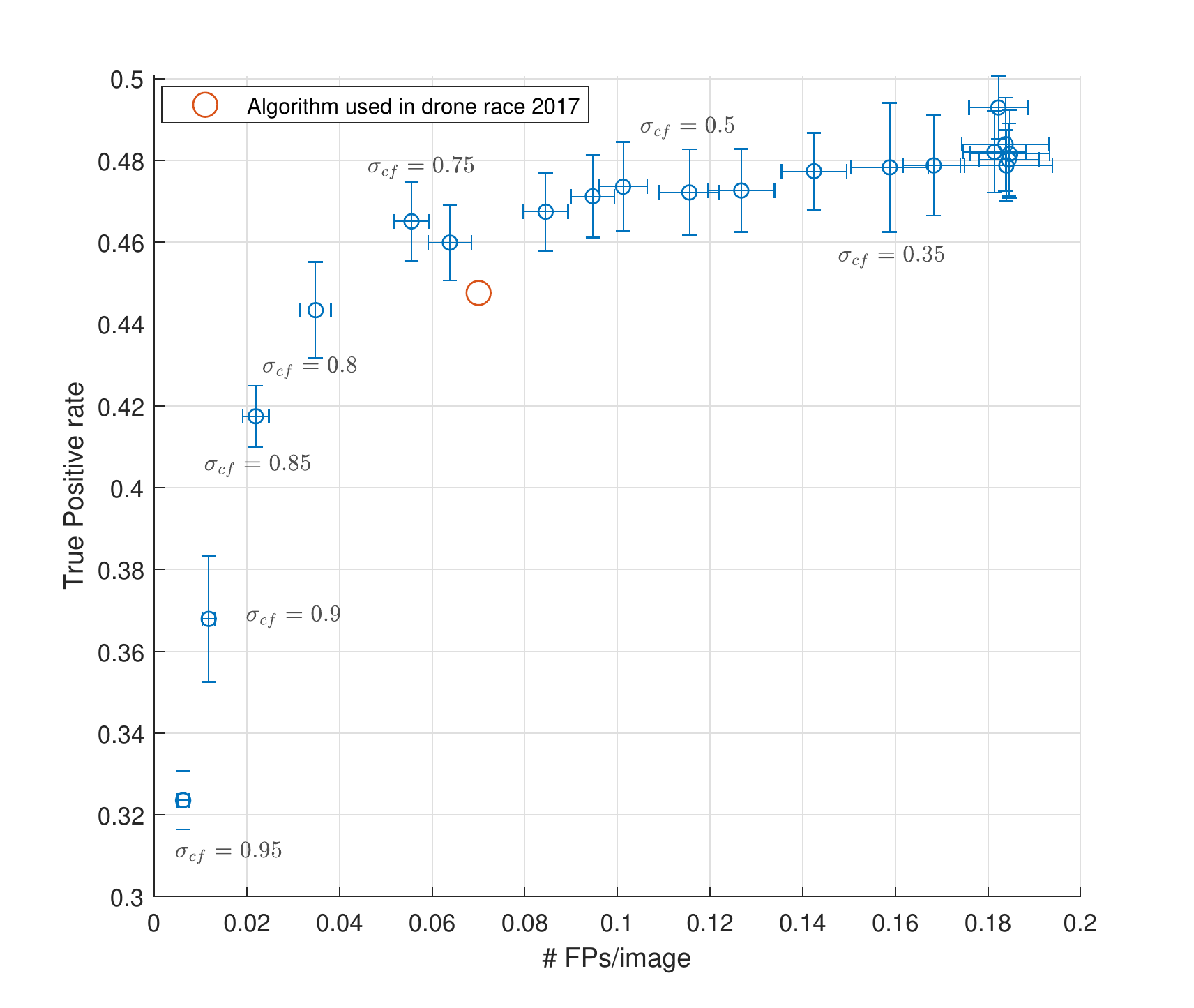}
    \caption{ROC curve with $\sigma_L=25$ and varying $\sigma_{cf}$}
    \label{fig:ROC cf}
\end{figure}

To evaluate the performance of the snake gate detection algorithm, 600 onboard images with/without gate are used to test the algorithm.(Figure \ref{fig:ROC examples}) The ROC curve with varying $\sigma_L$ is shown in Figure \ref{fig:ROC min length}. It should be noted that the detection is done 10 times with one $\sigma_L$ to obtain the statistical result. The x-axis of ROC curve is average of false positive detection per image and the y-axis is true positive rate. To make the trend in Figure \ref{fig:original ROC} clearer, we enlarge local part of the ROC curve by using logarithm coordinate system in Figure \ref{fig:local ROC}. From ROC curve, it can be seen that when $\sigma_L$ is small ($\sigma_L < 15$), the number of the false positive detections decreases significantly while $\sigma_L$ increases without sacrificing TPR.That is because $\sigma_L$ helps to reject the small detections caused by small color blocks of the environment. When $\sigma_L>35$, however, TPR decreases sharply, the reason is that $\sigma_L$ is too large to accept true positive detections. $\sigma_L = 25$ can give the optimal option with low FPs/image and almost highest TPR. Then, with $\sigma_L = 25$ , we draw another ROC curve with varying $\sigma_{cf}$, which is shown in Figure \ref{fig:ROC cf}. It can be seen that with increasing $\sigma_{cf}$, false positive detections decrease without significantly decreasing of TPR. 

In autonomous drone race 2017, we tuned $\sigma_L$ through experimental trial-and-error and accept the detection with highest color fitness, from which, the ROC point is plotted by red circle in Figure \ref{fig:ROC min length} and Figure \ref{fig:ROC cf}. It is remarkably close to the optimal thresholds one would pick, given this more extended analysis. Please note  that the algorithm used in the 2017 drone race only accepted the gate with the highest color fitness, and not every gate that was over the color threshold. It should also be noted that the method described above can be used for tuning $\sigma_L$ and $cf$ automatically. However, manually labeling and running snake gate detection on the dataset for each set of parameters is time-consuming especially when the drone needs to be deployed in a new racing track with limited preparing time.

\begin{figure}[!h]
    \centering
    \includegraphics[width=0.6\textwidth]{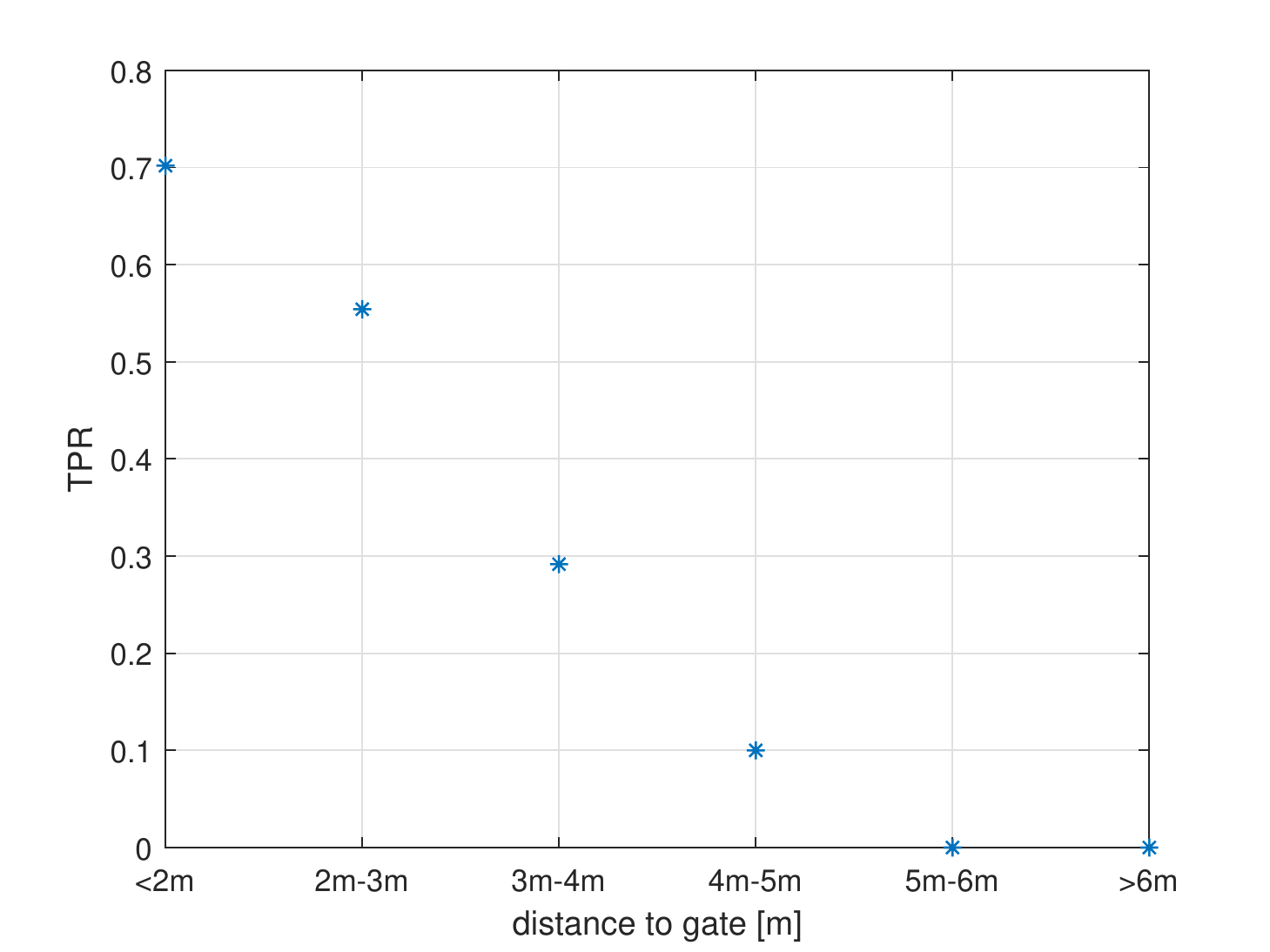}
    \caption{When the drone approaches the gate, true positive rate becomes larger because of larger and clearer gate on image plane}
    \label{fig:TPR vs distance}
\end{figure}

It should be noted that the true positive rate in above figures is the statistical result on the entire dataset. In order to evaluate how good or bad a true positive rate of 0.46 is, one has to take additional factors into account. Importantly, the distance between the drone and the gate can significantly affect the detection. Figure \ref{fig:TPR vs distance} shows how the true positive rate changes with the change of distance between the gate and the drone. It is very clear that when the drone gets closer to the gate, the snake gate detection has a higher true positive rate, reaching $70\%$ at close distances. 

\begin{figure}[!h]
    \centering
    \includegraphics[trim=10cm 0cm 10cm 0cm,width=0.8\textwidth]{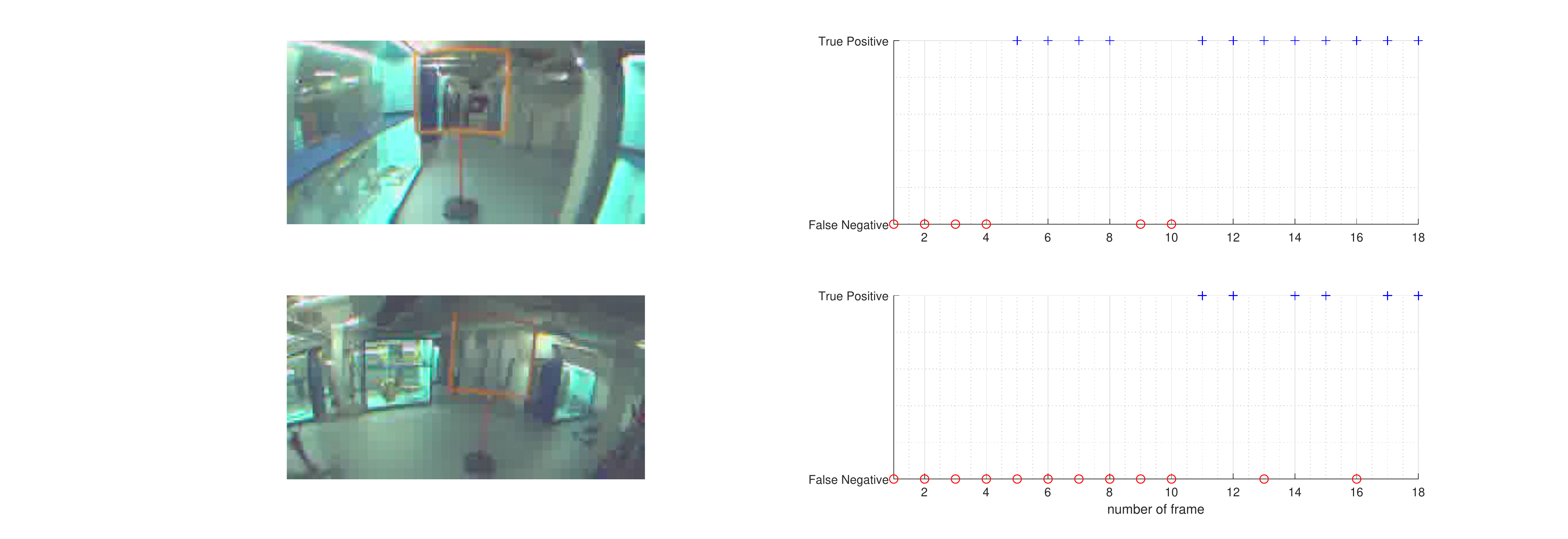}
    \caption{While the drone approaches the gate, there still exist some false negatives which may caused by light condition and distortion.}
    \label{fig:sequence}
\end{figure}

Figure \ref{fig:sequence} shows the detection result while the drone approaches the gate. In the beginning, the distance between the drone and the gate is large which leads to false negative detections. Once the drone starts detecting the gate, it can detect the gate most of times. However, there still exist some false negative detections. But these false negative detections could be handled by filters which will be explained in details next section. 

\begin{figure}[!h]
    \centering
    \includegraphics[width=0.5\textwidth]{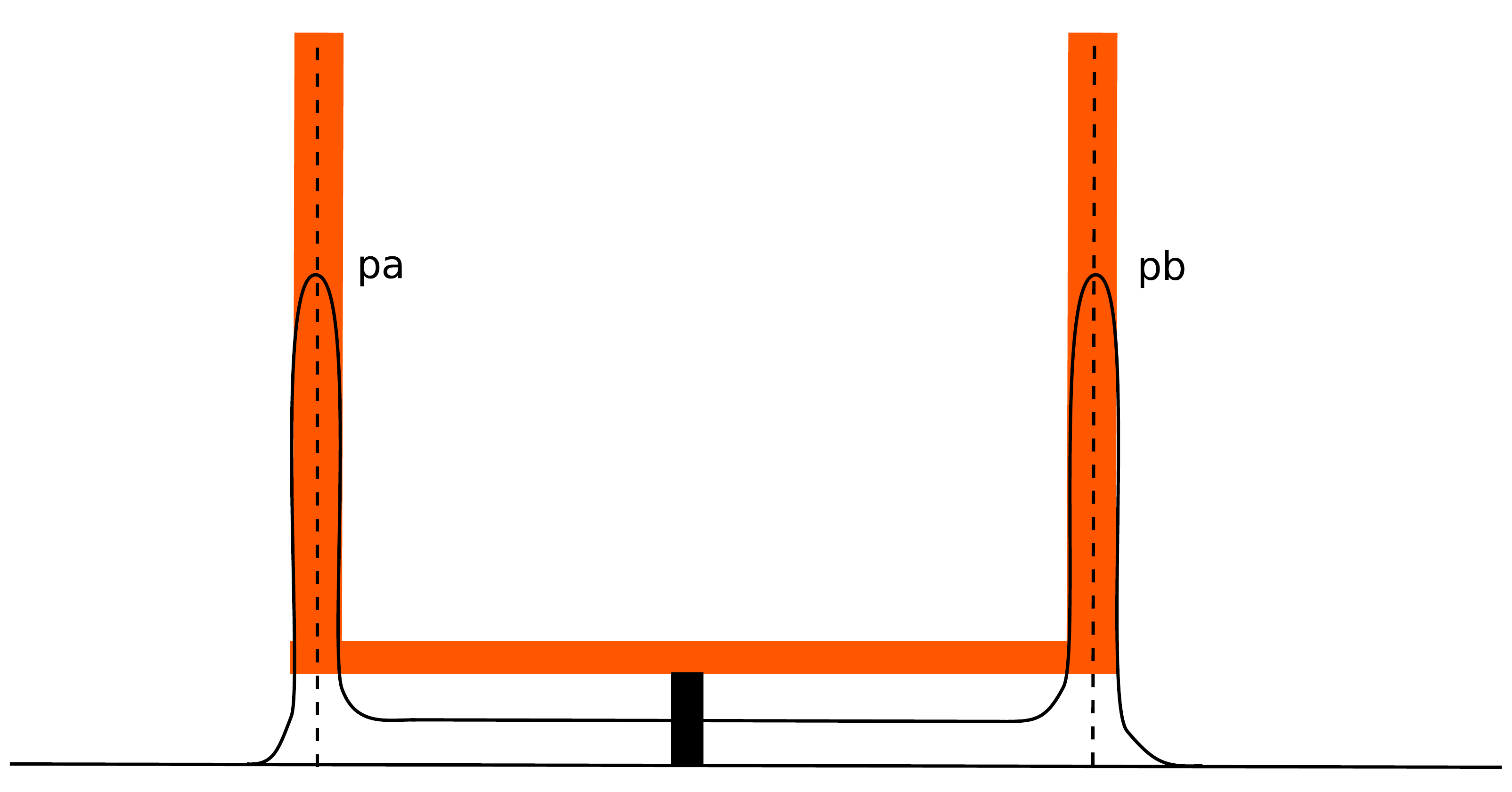}
    \caption{When the drone is close to the gate, only a part of the gate can be seen. A histogram of target color in x axis is employed. Two side bars can be found by the two peaks of the histogram}
    \label{fig:histogram_detection}
\end{figure}

When the drone is close to the gate ($<1m$), only part of the gate can be seen. In this scenario, snake gate detection will not detect the gate. A second detection called histogram gate side detection is employed to replace snake gate detection when the position estimate from the Kalman filter is $<$ 1m. (Figure \ref{fig:histogram_detection}) This detection algorithm accumulates the number of target color pixels by each column. Then two peaks of the histogram which represent two sidebars of the gate can be found. Later, the position of these two bars can be used by pose estimation to extract relative position between the gate and the drone. 

\subsection{Pose estimation}
When a gate with known geometry is detected, its image can provide the pose information of the drone. The problem of determining the position and orientation of a camera given its intrinsic parameters and a set of $n$ correspondences between 3D points and their 2D projections is called Perspective-n-Point (PnP) problem. \cite{bujnak2008general} In our case, 4 coplanar control points (gate corners) are available which leads to a unique solution.\cite{abidi1995new} However, PnP is sensitive to the mismatches of 3D points and 2D points which in our case is inevitable because the vibration and complex environment. Therefore, these methods are usually combined with RANSAC scheme to reject noise and outliers. Unfortunately, the fact that only four corner points are available on one gate limits the effectiveness of such a scheme. In this section, a novel algorithm combining gate detection result and the onboard AHRS attitude estimation will be derived to provide the pose estimation of the drone.

\begin{figure}[!h]
    \centering
    \includegraphics[width=0.5\textwidth]{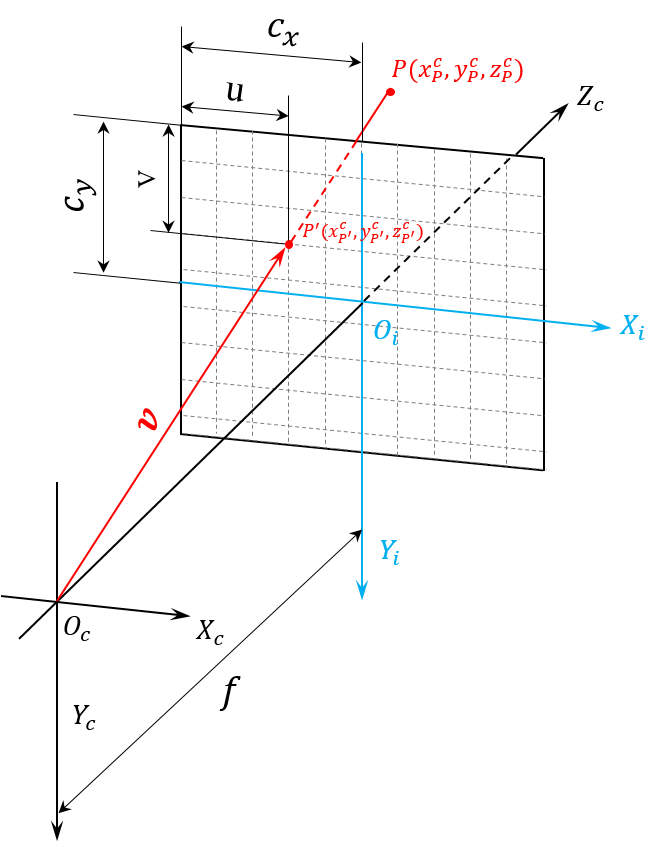}
    \caption{A pinhole camera model. $O_c$ is the focal point and the origin of camera frame $O_cX_cY_cZ_c$. $f$ is the focus. $O_iX_iY_iZ_i$ is image frame. $\mathbf{P}$ is a 3D point in space and $\mathbf{P}^{'}$ is its image point on image plane} 
    \label{fig:pinholeModel}
\end{figure}

Since we are using a fish-eye camera, a calibration procedure should be done first \cite{dhane2012generic}. Then, the camera can be simplified as a pinhole camera model. (Figure \ref{fig:pinholeModel}) According to the similar triangle principle, we have

\begin{align}
\begin{bmatrix}
x_{\mathbf{P}^{'}}^C \\ y_{\mathbf{P}^{'}}^C
\end{bmatrix} = 
\begin{bmatrix}
f & 0 \\ 0 & f
\end{bmatrix}
\begin{bmatrix}
\frac{x_{\mathbf{P}}^C}{z_{\mathbf{P}}^C} \\ \frac{y_{\mathbf{P}}^C}{z_{\mathbf{P}}^C}
\end{bmatrix}
\label{equ:pinhole model}
\end{align}

Assume that each pixel's size is $d_x$ and length $d_y$ and the principle points' coordinate is $(C_x,C_y)$, we could transfer the pinhole model \ref{equ:pinhole model} to 

\begin{align}
\begin{bmatrix}
u \\ v
\end{bmatrix}= 
\begin{bmatrix}
\frac{f}{d_x} & 0 \\ 0 & \frac{f}{d_y}
\end{bmatrix}
\begin{bmatrix}
\frac{x_{\mathbf{P}}^c}{z_{\mathbf{P}}^c} \\ \frac{y_{\mathbf{P}}^c}{z_{\mathbf{P}}^c}
\end{bmatrix} + 
\begin{bmatrix}
C_x \\ C_y
\end{bmatrix}
\label{equ:pinholeModel_pixel}
\end{align}

To write the pinhole model \ref{equ:pinholeModel_pixel} in homogeneous coordinates, we have

\begin{align}
\begin{bmatrix}
u \\ v \\ 1
\end{bmatrix} = 
\begin{bmatrix}
f_x & 0 & C_x \\ 0 & f_y & C_y  \\ 0 & 0 & 1
\end{bmatrix}
\begin{bmatrix}
\frac{x_{\mathbf{P}}^c}{z_{\mathbf{P}}^c} \\ \frac{y_{\mathbf{P}}^c}{z_{\mathbf{P}}^c} \\ 1
\end{bmatrix}
\end{align}

where $f_x=f/d_x$, $f_y=f/d_y$. $u$, $v$,$C_x$ and $C_y$ are in pixel unit. From Figure \ref{fig:pinholeModel}, it can be seen that the 3D point $P$, the image point $P^{'}$ and the focal point $O_c$ are on one line. Thus, the direction of the light ray from $O_c$ to $P$ can be described by a bearing vector $\mathbf{v}$ which can be expressed in camera frame by 
\begin{align}
\mathbf{v} = 
\begin{bmatrix}
v_x^c \\ v_y^c \\ v_z^c 
\end{bmatrix} =
\begin{bmatrix}
(u-C_x)/f_x \\ (v-C_y)/f_y \\ 1
\end{bmatrix}
\end{align}

To express vector $\mathbf{v}$ in earth frame, we introduce 2 rotation matrices $\Re_C^B$ and $\Re_B^E$. $\Re_C^B$ is the rotation matrix from camera frame $C$ to body frame $B$ which is a fixed matrix. $\Re_C^B$ is the rotation matrix from body frame $B$ to earth frame $E$ consist of three Euler angle $\psi$, $\theta$ and $\phi$, which can be measured from onboard AHRS system. Thus, bearing vector $\mathbf{v}$ could be expressed in the Earth frame $E$ by

\begin{align}
\mathbf{v} = 
\begin{bmatrix}
v_x^E \\ v_y^E \\ v_z^E
\end{bmatrix}=\Re_B^E\Re_C^B
\begin{bmatrix}
v_x^c \\ v_y^c \\ v_z^c 
\end{bmatrix}
\end{align}

A line passing through point $\mathbf{P}$ with direction $\mathbf{v}$ can be written as 
\begin{align}
L(\mathbf{p},\mathbf{v}) = \mathbf{p} + \lambda\mathbf{v},{}{} \lambda\in[-\infty,+\infty]
\end{align}

The perpendicular distance $D(\mathbf{t};\mathbf{p},\mathbf{v})$ of a point $\mathbf{t}$ to line $L(\mathbf{p},\mathbf{v})$ is

\begin{align}
D(\mathbf{t};\mathbf{p},\mathbf{v}) = \norm{(\mathbf{p}-\mathbf{t})-((\mathbf{p}-\mathbf{t})^{T}\mathbf{v})\mathbf{v}}_2
\end{align}

\begin{figure}[!h]
\centering
 \includegraphics[scale=.5]{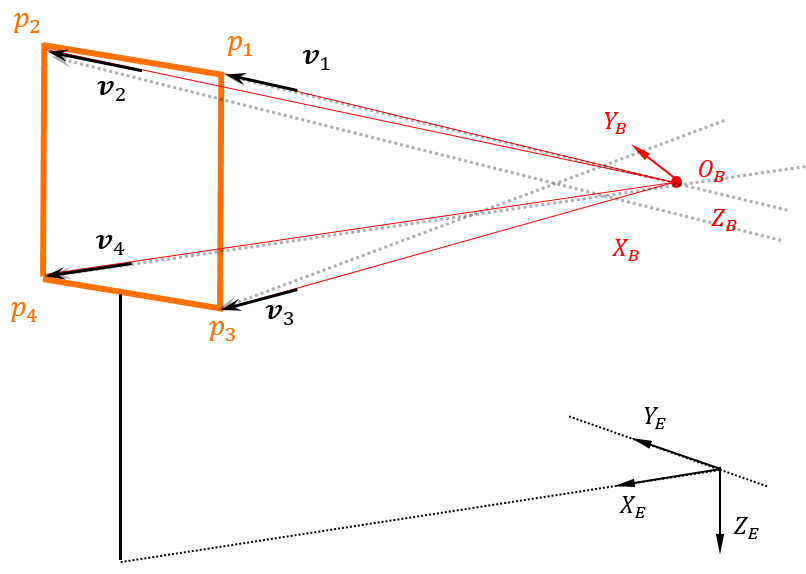}
 \caption{Four light rays from four corners of the gate with bearing vector $\mathbf{v}_i$, which could be calculated by four corner's images on image plane and camera's intrinsic parameters, should intersect at focal points. (red line) However, wrong bearing vectors from wrong detections could make the light rays not intersect at one point.} 
    \label{fig:wrong detections}
\end{figure}

According to the pinhole model, 4 light rays with bearing vectors $\mathbf{v}_i$ from four corners  of the gate should intersect at the focal point $\mathbf{t}$ ( Figure \ref{fig:wrong detections}), which is the position of the drone. The bearing vectors can be calculated by the four points' images on the image plane and camera's intrinsic parameters. This intersection point could be calculated analytically. However, due to the detection error of the gate's corners, bearing vectors can be wrongly calculated, for example, in Figure \ref{fig:wrong detections} four light rays do not intersect at one point.(gray line) Thus, there is no analytical solution of camera's position. Instead of finding analytical solution of camera's position, a numerical solution is found that finds a point whose distance to the four light rays is minimum. Hence, estimating the position of the drone can be converted to an optimization problem that finds an optimal point $\mathbf{t}$ which has minimal distance to 4 light rays, which can be expressed mathematically by 

\begin{align}
\min_{\mathbf{t}}\sum_{i=1}^4D(\mathbf{t};\mathbf{p}_i,\mathbf{v}_i)
\end{align}

which is a least squares problem. 

\begin{figure}[!h]
\centering
 \includegraphics[scale=.5]{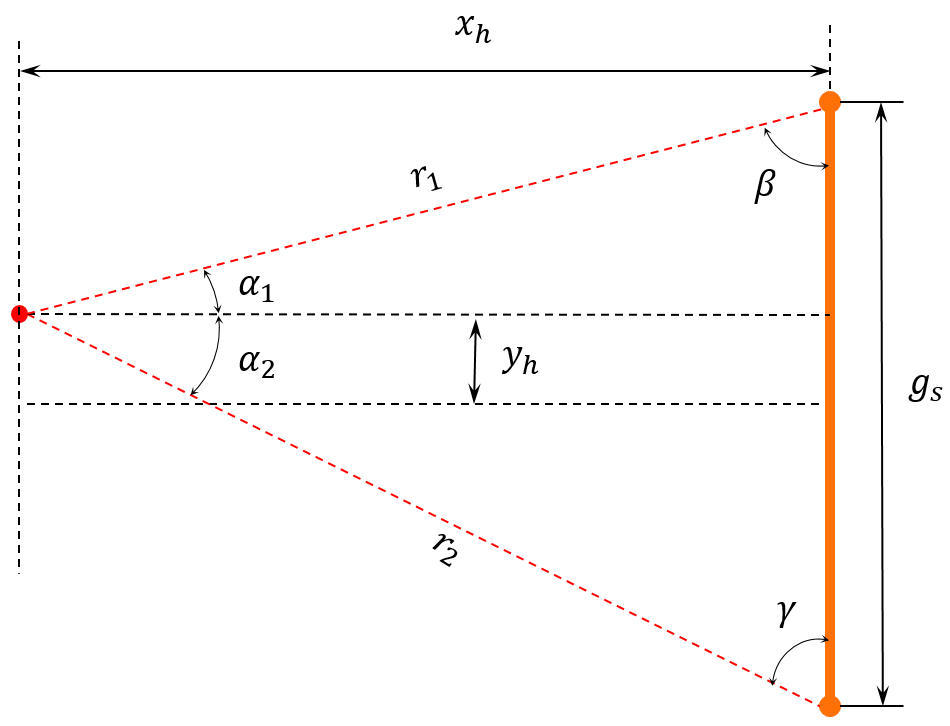}
 \caption{The top view of the position of the drone and the gate} 
    \label{fig:histogram estimation}
\end{figure}

When the drone is close to the gate, only two sidebars can be detected by the histogram method. With the position of bars on the image plane, the pose of the drone can be estimated by geometrical principle. In Figure \ref{fig:histogram estimation}, $\alpha_1$ and $\alpha_2$ are calculated by the position of the image of two bars on image plane and intrinsic parameters. Then we have

\begin{align}
\begin{split}
\gamma =& \frac{\pi}{2}-\alpha_2 \\
\frac{r_1}{\sin\gamma} =& \frac{g_s}{\sin(\alpha_1+\alpha_2)} \\
x_h =& r_1\cos\alpha_1 \\
y_h =& \frac{g_s}{2}-r_1\sin\alpha_1
\end{split}
\end{align}

where $g_s$ is the length of the gate. Hence, based on the detection of the histogram peaks in the image (corresponding to $\alpha_1$ and $\alpha_2$), we can deduce the lateral position of the camera with respect to the gate ($x_h$ and $y_h$).

\begin{figure}[h!]
\centering
\begin{tabular}{c c c}
\subfloat[]
{
  \includegraphics[scale=.45]{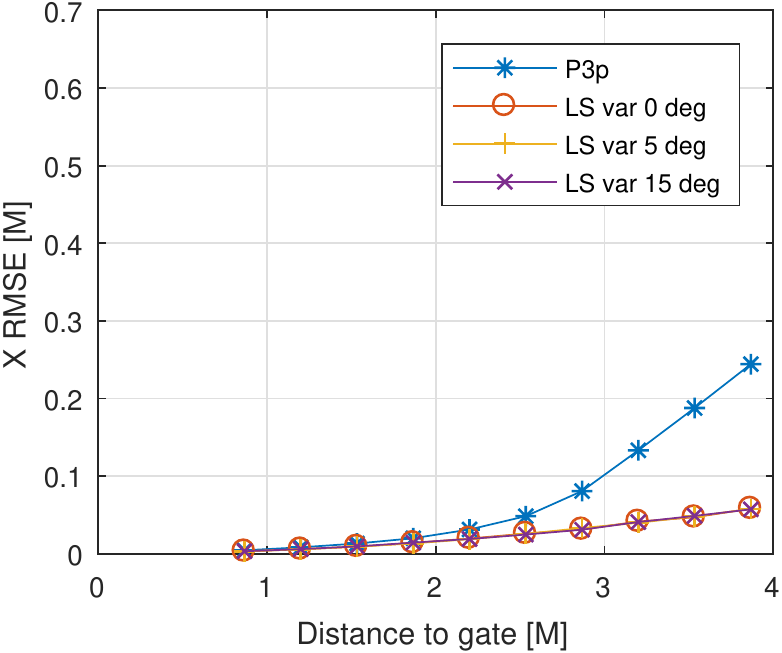}
}\hfill

\subfloat[]
{
  \includegraphics[scale=.45]{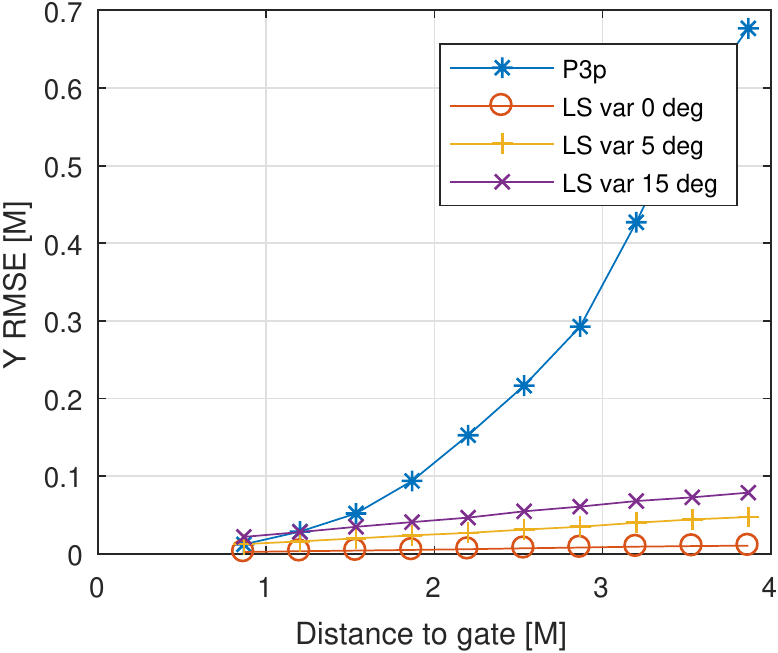}
  \label{fig:overexposure}
}
\hfill
\subfloat[]
{
  \includegraphics[scale=.45]{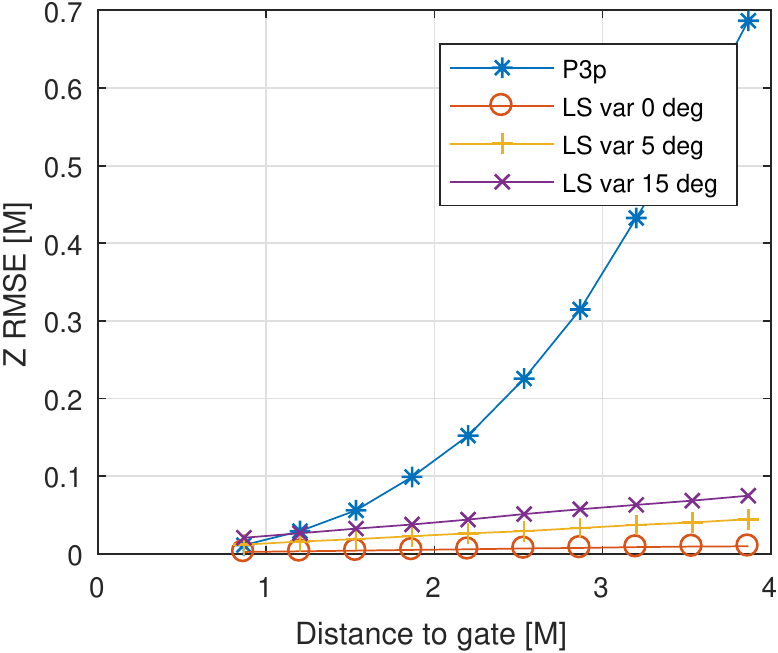}
  \label{fig:overexposure}
}
\end{tabular}
\caption{Simulation result of P3P and LS method. With the incrementation of the distance between the drone and the gate, both methods' error increase. However, LS method has much less error than P3P.}
\label{fig:LS vs PnP}
\end{figure}

A simulation is done to test the performance of our algorithm and compare it with a standard PnP method. For simulation, artificial gates are created, which are projected onto a virtual pinhole camera image. Since gate detections contain image noise and outliers, a set of real gate detections are compared with ground truth data. Based on this test the vision method experiments will therefore contain image noise with a standard deviation of 3.5 pixels. The Root Mean Squared Error (RMSE) is used to evaluate the performance of both algorithms. The result is shown in Figure \ref{fig:LS vs PnP} where each point represents a thousand trails of the position estimation algorithm in the presence of pixel noise. It can be seen that the error varies mainly as a function of distance to the gate. The LS method uses prior knowledge of the attitude and heading of the vehicle to obtain a more accurate position estimate. To study the effect of attitude error, noise with a variance of 0, 5 and 15 degrees is added to the attitude and heading estimates. It is clear from the figure that the LS method has far higher accuracy in RMSE compared to the PnP method, even in the presence of relatively large noise in the attitude estimate.

\begin{figure}[h!]
\centering
\begin{tabular}{c c}
\subfloat[]
{
  \includegraphics[scale=.6]{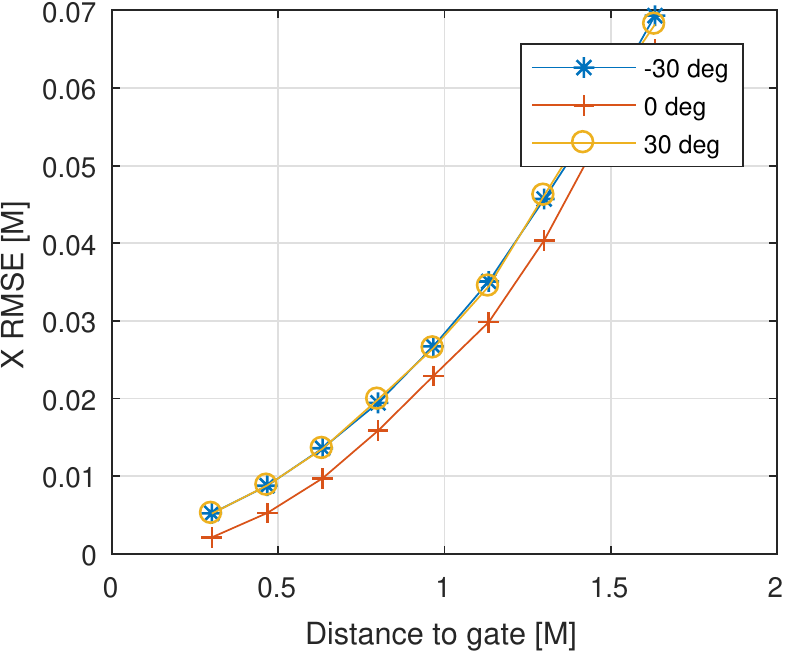}
}\hfill
&
\subfloat[]
{
  \includegraphics[scale=.6]{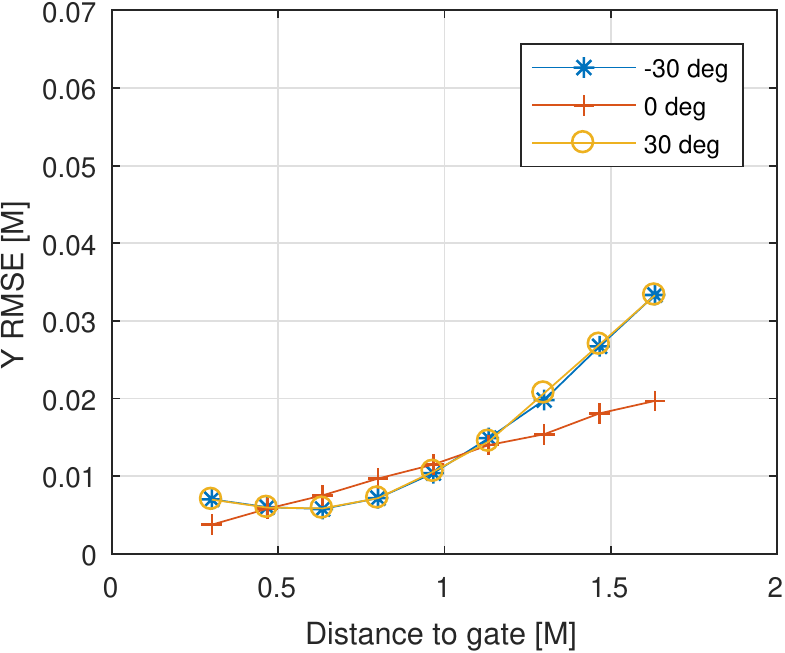}
  \label{fig:overexposure}
}
\end{tabular}
\caption{X and Y histogram position RMSE as function of distance to the gate}
\label{fig:histogram}
\end{figure}

Also, the histogram position estimation method is evaluated in simulation. Similar to the LS method, pixel noise with a standard deviation of 3.5 is introduced. Figure \ref{fig:histogram} shows the results of the position RMSE in the horizontal plane in x and y-direction. The experiment is performed with a heading angle of -30, 0 and 30 degrees. From the figure, it can be observed that the position error of this method is relatively low. However, in reality, the method is only effective up to a maximum distance of 1.5 meters, due to the possible background color leading to spurious histogram peaks that are hard to filter out.

\subsection{Vision-IMU state estimation}
In order to close the control loop, state estimation is essential since the measurements (in our case, distance from vision, acceleration and angular velocity from IMU) are inevitably noisy and biased. A common approach for state estimation is the Kalman filter and its variants such as extended Kalman filter(EKF), Unscented Kalman Filter(UKF) and Particle filtering. In the field of UAVs, 15-states (position $\mathbf{x}$, velocity $\mathbf{v}$, attitude $\mathbf{\phi}$ and IMU bias $\mathbf{b}$) Kalman filter is used commonly in many scenarios. It first integrates angular rate to gain rotation matrix from body to earth $\Re_B^E$. Next, $\Re_B^E$ is used to rotate acceleration measured by the accelerometer to earth frame. Then, the acceleration will be integrated twice to gain the position. And finally, position measurement will be used to correct the position prediction. Usually, UAVs' onboard IMUs are low-cost MEMS which suffer from biases and noise severely. During the prediction phase, the bias of accelerometer is integrated twice which may cause the prediction to deviate from the real position over time. If the position measurement has a relatively high frequency, the deviation of the position prediction could be corrected before it diverges. At the same time, the bias of IMU could also be estimated as states in the system and it should converge in short time. However, in our case, position measurements come from onboard image processing which has a low rate of around 20 HZ and the drone may cover significant durations without vision measurements. In this case, position prediction may deviate largely before new position measurement comes. Thus, the bias estimation converges slowly. In this section, we adopt the drone's aerodynamics model to the prediction model in Kalman filter which has a better performance than classic 15-states Kalman filter. 

The kinematics of the drone can be described by

\begin{align}
\dot{\mathbf{X}} = \mathbf{V}
\end{align}

To express $\mathbf{V}$ in body frame, we have

\begin{align}
\begin{bmatrix}
\dot{x}^E \\ \dot{y}^E  \\ \dot{z}^E
\end{bmatrix}=\Re_B^E(\phi,\theta,\psi)
\begin{bmatrix}
v_x^B \\ v_y^B \\v_z^B
\end{bmatrix}
\label{equ:kinematic}
\end{align}

where,$x^E,y^E,z^E$ are the drone's position in earth frame $E$. $v_x^B, v_y^B, v_z^B$ are the drone's velocity in body frame $B$. One property of the onboard accelerometer is that it measures specific force $\mathbf{F}_s$ in body frame $B$ instead of vehicle's acceleration. The specific force in $\mathbf{Z}^B$ direction is mainly caused by thrust $\mathbf{T}$ under the assumption that the thrust of quadrotor is aligned with $\mathbf{Z}^B$. The force acting on $\mathbf{X}^B$ and $\mathbf{Y}^B$ can be caused by many factors, for instance, blade flapping, profile drag, and translational drag. But they could be approximated as a linear function, assuming that the indoor environment has no wind: \cite{svacha2017improving}

\begin{align}
\begin{bmatrix}
a_x^B \\ a_y^B
\end{bmatrix} = 
\begin{bmatrix}
k_x & 0 \\ 0 & k_y
\end{bmatrix}
\begin{bmatrix}
v_x^B \\ v_y^B
\end{bmatrix}
\end{align}
where $k_x$ and $k_y$ are drag coefficient which could be identified off-line. With this property, the accelerometer can actually provide the information of velocity of the drone by

\begin{align}
\begin{bmatrix}
v_x^B \\ v_y^B
\end{bmatrix} = 
\begin{bmatrix}
k_x & 0 \\ 0 & k_y
\end{bmatrix}^{-1}
\begin{bmatrix}
a_x^m-b_a^x \\ a_y^m-b_a^y
\end{bmatrix}
\label{equ:acc 2 vel}
\end{align}
where $a_x^m$ and $a_y^m$ are the measurement of accelerometer. $b_a^x$ and $b_a^y$ are the bias of accelerometer. Combine equation \ref{equ:kinematic} and equation \ref{equ:acc 2 vel}, we have 

\begin{align}
\begin{bmatrix}
\dot{x}^E \\ \dot{y}^E  \\ \dot{z}^E
\end{bmatrix}=\Re_B^E(\phi,\theta,\psi)
\begin{bmatrix}
k_x & 0 & 0 \\ 0 & k_y & 0 \\ 0 & 0 & 1
\end{bmatrix}^{-1}
\begin{bmatrix}
a_x^m-b_a^x \\ a_y^m-b_a^y \\ v_z^B
\end{bmatrix}
\label{equ:kinematic + acc}
\end{align}

In equation \ref{equ:kinematic + acc}, the bias only needs to be integrated once to predict the position of the drone instead of being integrated twice in original 15-states Kalman filter, which could help to decrease the error of prediction.

As mentioned above, the onboard AHRS system is a complementary filter, which on a low level fuses accelerometer and gyro data to estimate the attitude of the drone. It can directly provide the attitude estimation to the outer loop. The AHRS fusing only IMU data may introduce a bias to the attitude estimation. In this paper, we assume that this low level attitude estimation bias can be neglected. Hence, AHRS and accelerometer reading can be used as inputs to propagate the prediction model \ref{equ:kinematic + acc}. 

According to Newton's laws of motion, the motion of the drone can be described as

\begin{align}
\begin{bmatrix}
\dot{v}_x^B \\ \dot{v}_y^B \\ \dot{v}_z^B
\end{bmatrix}=
\Re_E^B \begin{bmatrix}
0 \\ 0 \\ g
\end{bmatrix}+\begin{bmatrix}
a_x^m-b_a^x \\ a_y^m-b_a^y \\ a_z^m-b_a^z
\end{bmatrix}-\begin{bmatrix}
p \\ q \\ r
\end{bmatrix}\times\begin{bmatrix}
v_x^B \\ v_y^B \\ v_z^B
\end{bmatrix}
\label{equ:velocity dot}
\end{align}

where $g$ is gravity factor and $p,q,r$ are angular velocity in body frame $B$ measured by the gyro. Since in equation \ref{equ:kinematic + acc}, body velocity has already had measurements from the accelerometer, in equation \ref{equ:velocity dot}, we omit the first 2 equations and only leave the last equation combining with \ref{equ:acc 2 vel}, which results 
\begin{align}
\dot{v}_z^B = a_z^m-b_a^z+g\cos\theta\cos\phi+q\frac{a_x^m-b_a^x}{k_x}-p\frac{a_y^m-b_a^y}{k_y}
\label{equ:v_z dot}
\end{align}

With the assumption that gyro's bias is small, which can be neglected and the accelerometer's bias changes slowly,

\begin{align}
\dot{\mathbf{b}}_a=
\begin{bmatrix}
\dot{b}_a^x \\ \dot{b}_a^y \\ \dot{b}_a^z
\end{bmatrix}=\begin{bmatrix}
0 \\ 0 \\ 0
\end{bmatrix}
\label{equ:bias model}
\end{align}

Combining equation \ref{equ:kinematic + acc}, equation \ref{equ:v_z dot} and equation \ref{equ:bias model}, we have the process model for EKF as:

\begin{align}
\dot{\mathbf{x}}=\mathbf{f}(\mathbf{x},\mathbf{u})
\label{equ:EKF model}
\end{align}

with states and inputs defined by

\begin{align}
\mathbf{x}&=[x_E,y_E,z_E,v_z^B,{b}_a^x,{b}_a^y,{b}_a^z]^{T} \\ \mathbf{u}&=[\phi,\theta,\psi,a_x^m,a_y^m,a_z^m,p,q]^{T}
\end{align}

Then, a standard EKF predict/update procedure will be done to estimate the states, which can be found in Appendix.

\begin{figure}[h!]
\centering
\subfloat[]
{
  \includegraphics[scale=.6]{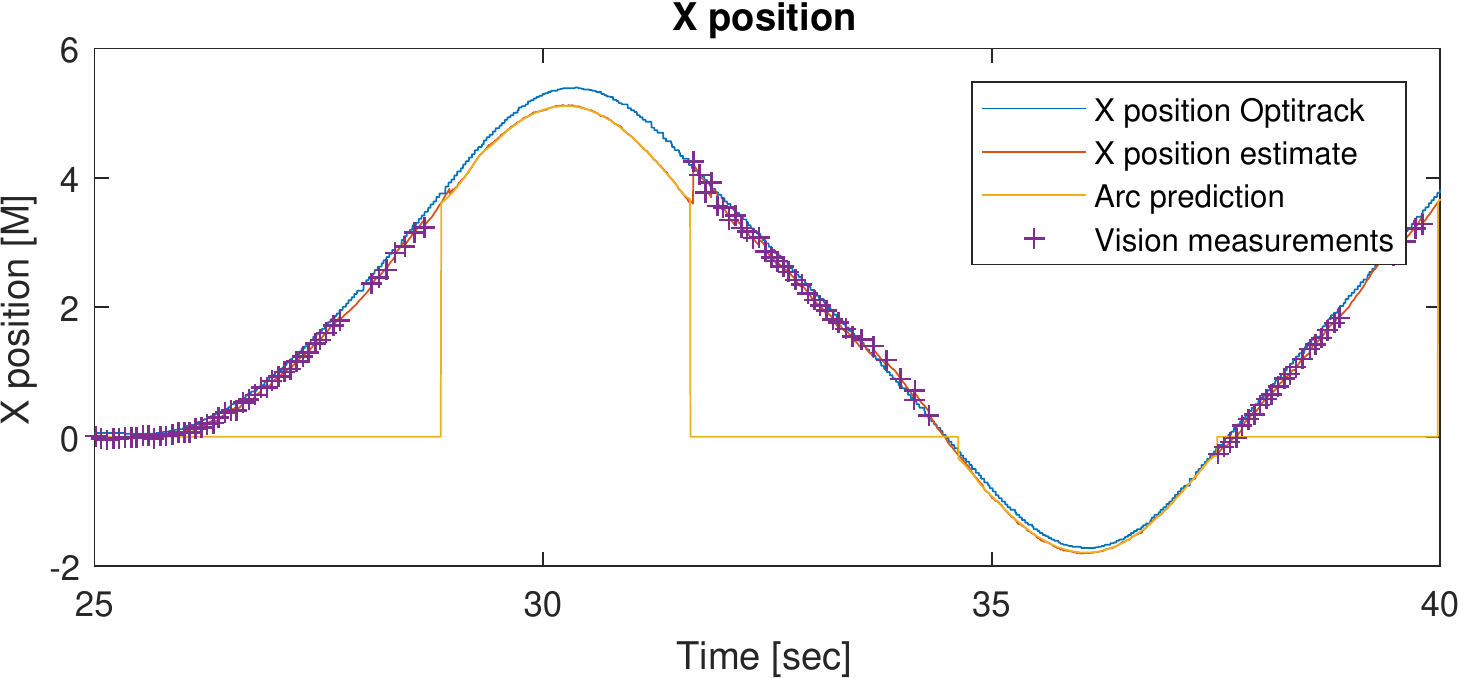}
}

\subfloat[]
{
  \includegraphics[scale=.6]{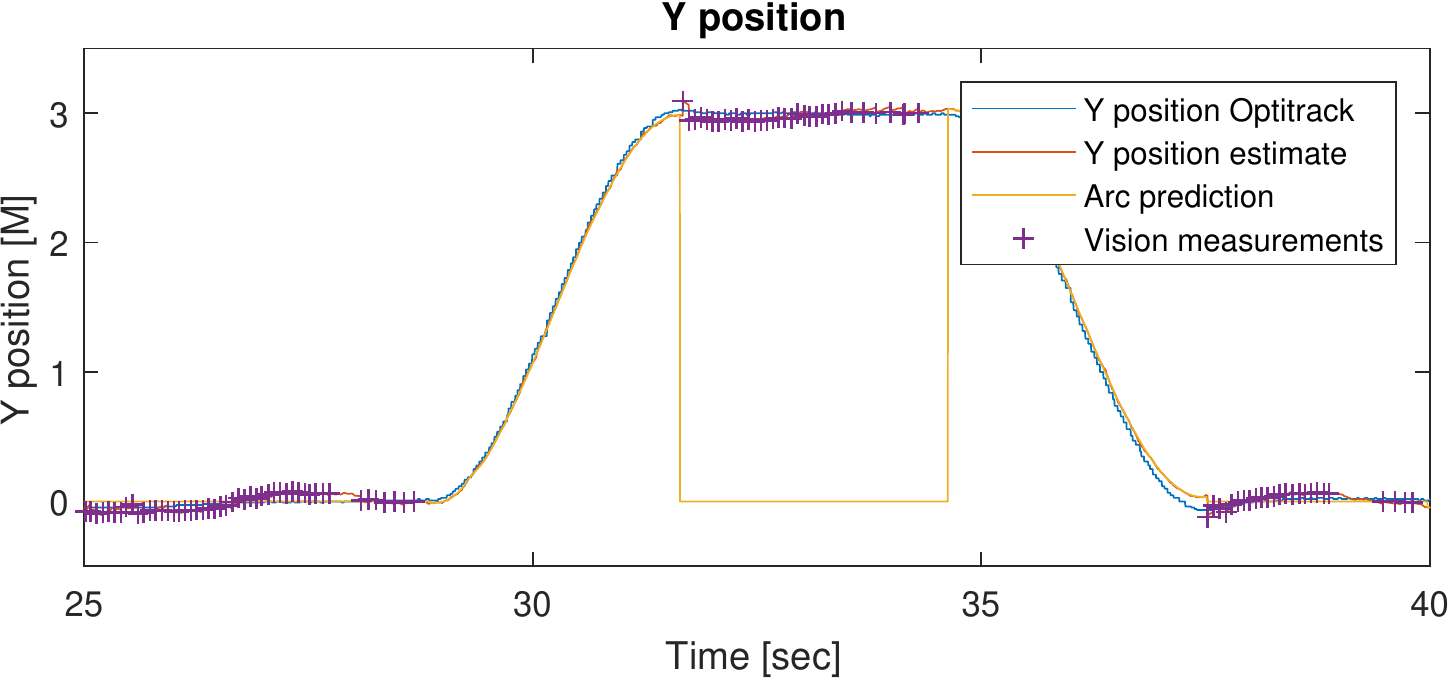}
  \label{fig:overexposure}
}
\hfill
\subfloat[]
{
  \includegraphics[scale=.6]{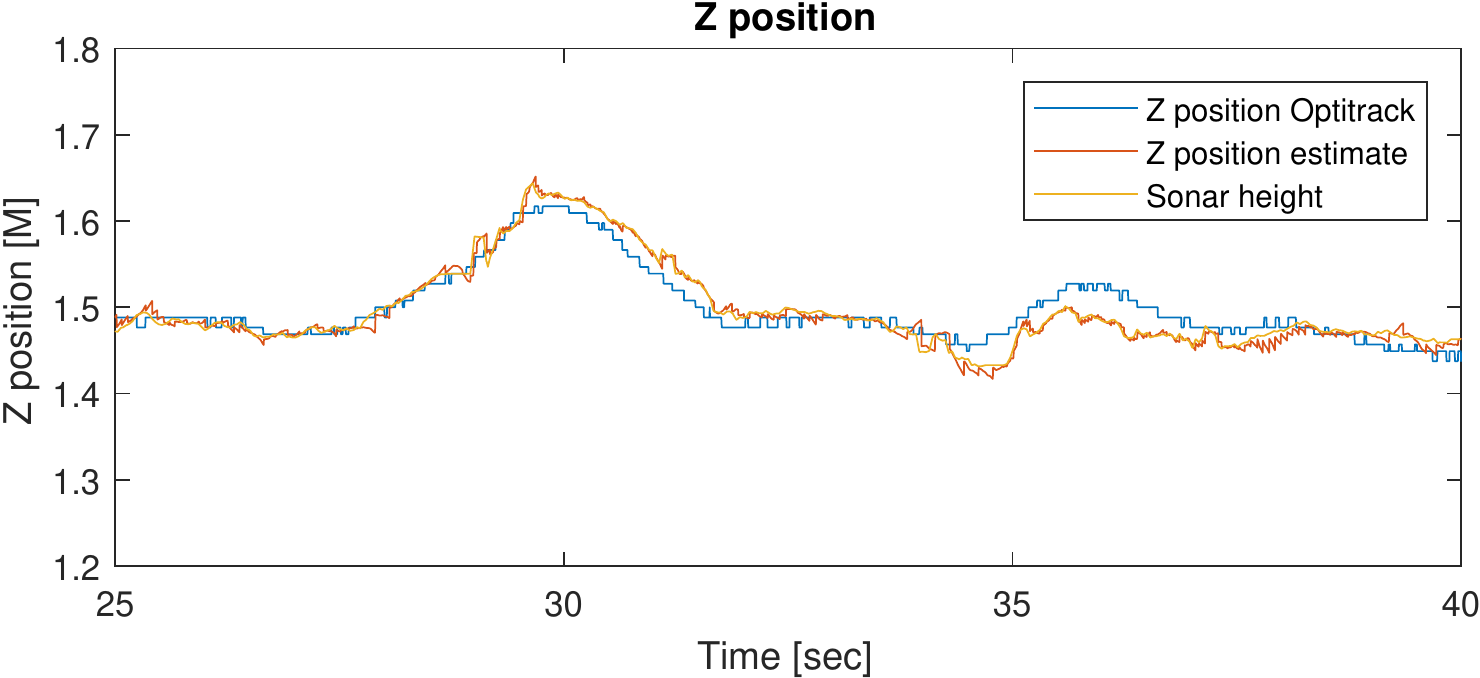}
  \label{fig:overexposure}
}
{
  \includegraphics[scale=.6]{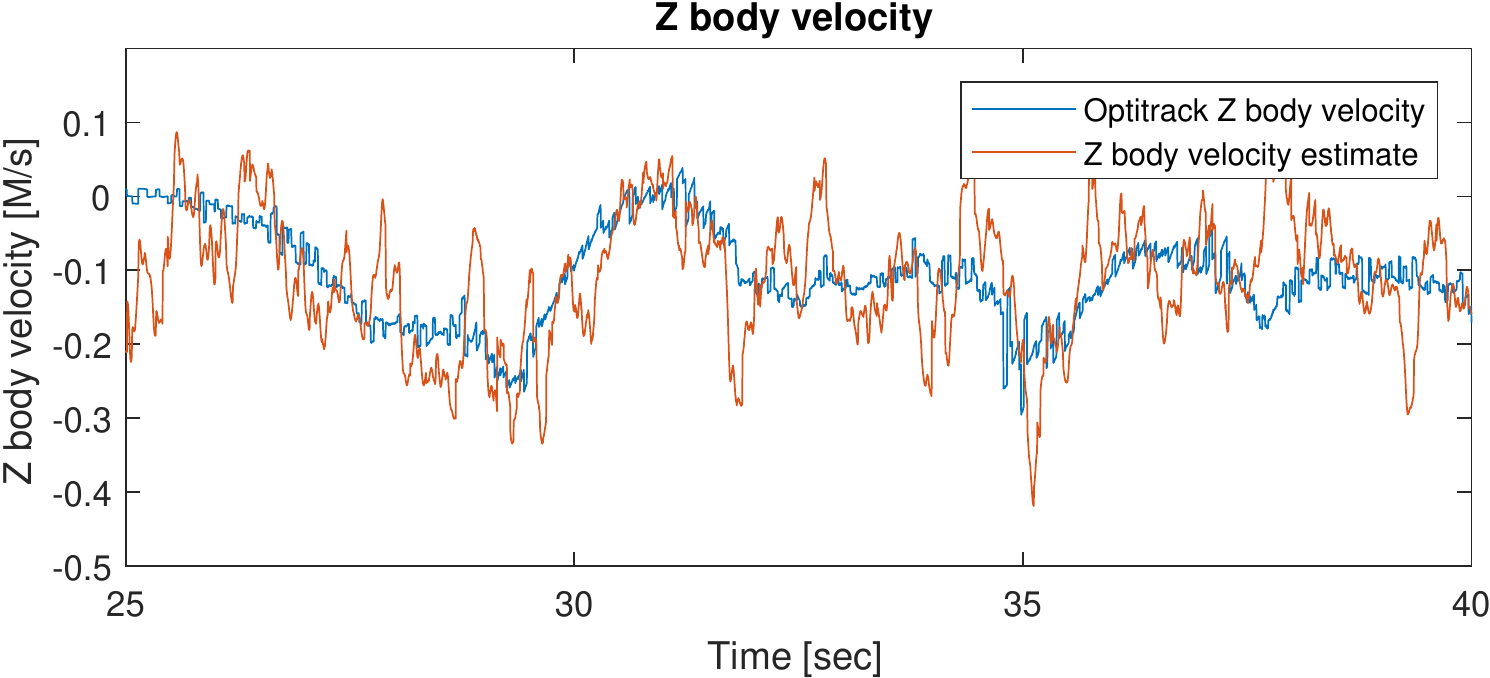}
  \label{fig:overexposure}
}
\caption{Extended Kalman filter result. The straight part flight is done with Opti-track. The vision pose estimation is done onboard. The arc part is done only by state prediction without the involvement of Opti-track.}
\label{fig:EKF result}
\end{figure}

\begin{figure}[h!]
\centering
\begin{tabular}{c c c}
\subfloat[]
{
  \includegraphics[scale=.6]{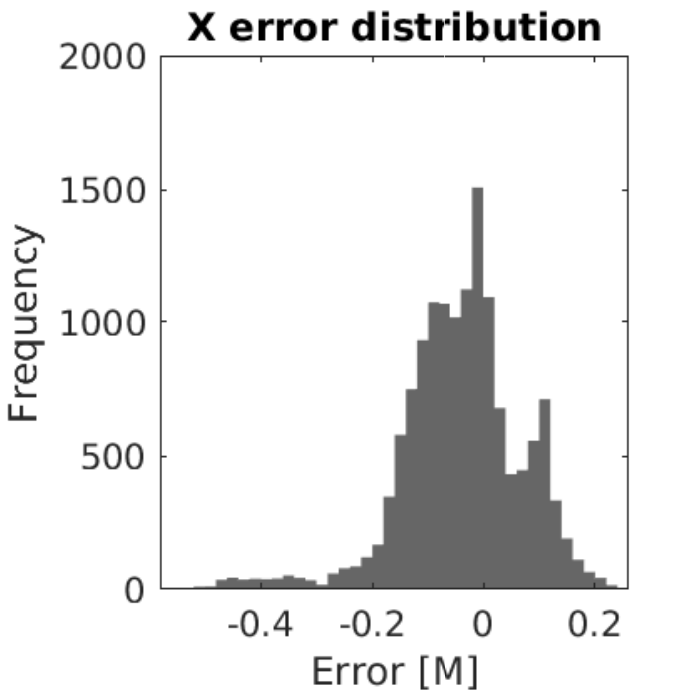}
}\hfill

\subfloat[]
{
  \includegraphics[scale=.6]{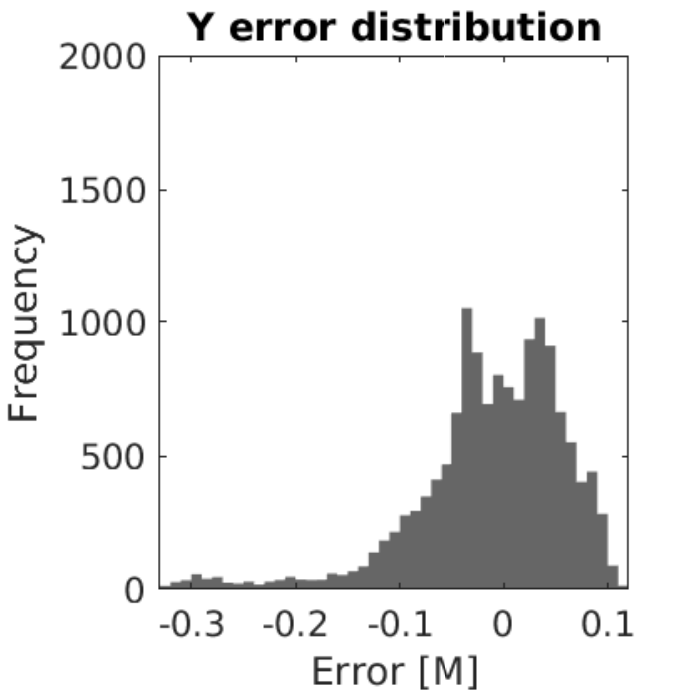}
  \label{fig:overexposure}
}
\hfill
\subfloat[]
{
  \includegraphics[scale=.6]{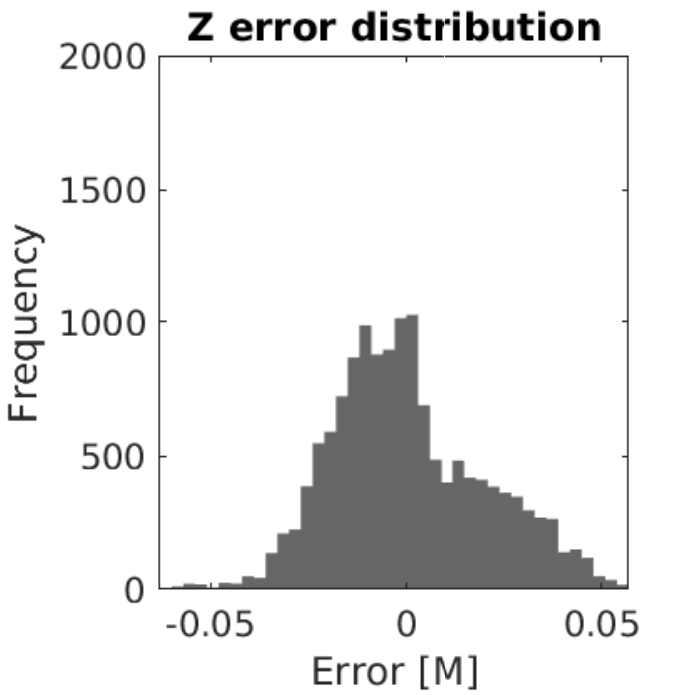}
  \label{fig:overexposure}
}
\end{tabular}
\caption{Extended Kalman filter error distribution}
\label{fig:EKF error}
\end{figure}

\begin{figure}[!h]
    \centering
    \includegraphics[width=\textwidth]{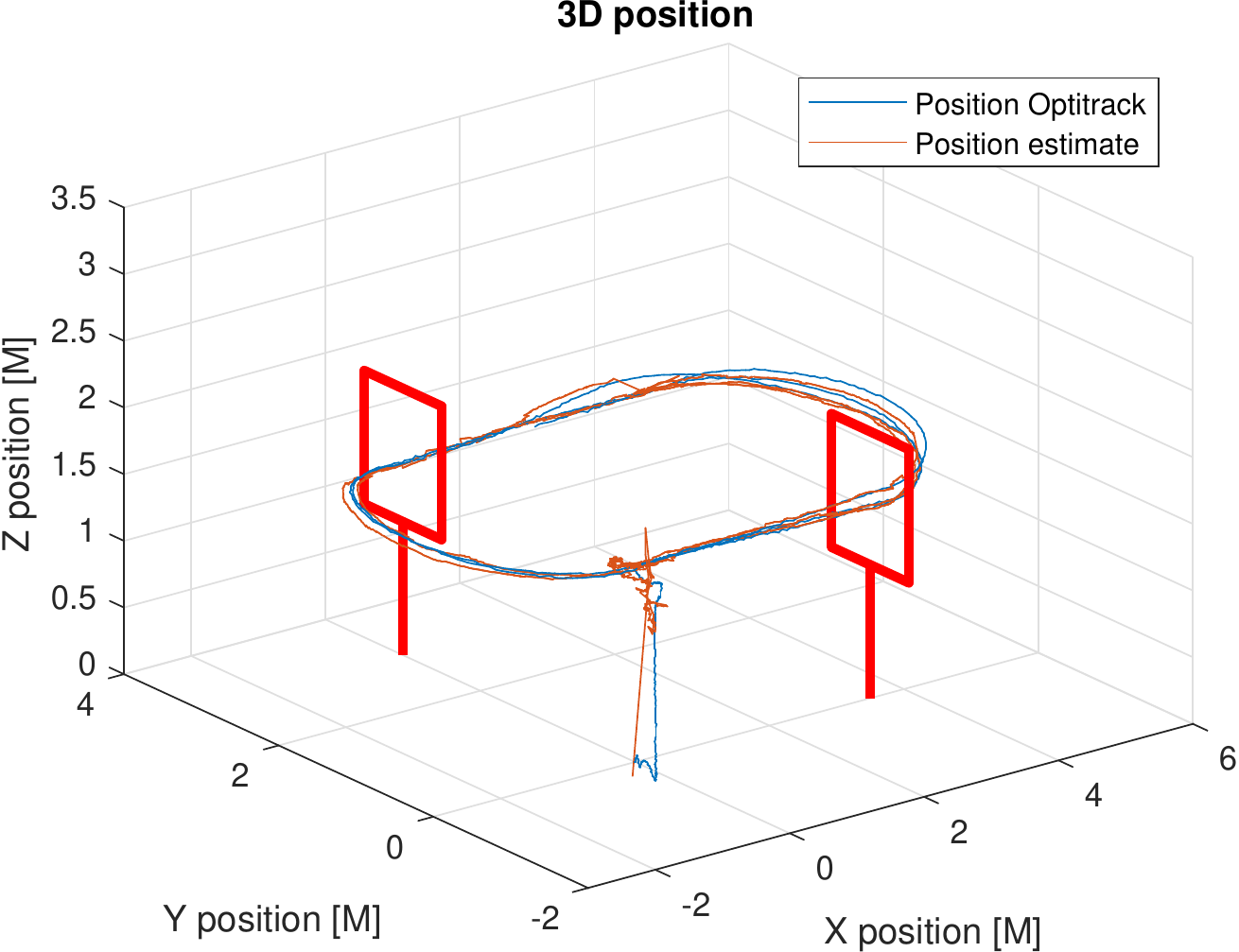}
    \caption{Experiment setup. The drone takes off from the ground and flies an oval cyclically. In straight parts, Opti-track is used to help the drone align with the gates, while vision detection is done onboard for logging. In the arc parts, a feed-forward control with state prediction is employed, which explains the reason the arcs end up at slightly different points.} 
    \label{fig:3D trajectory}
\end{figure}

To evaluate the performance of the visual navigation method described in this section, a flight test with a simplified two-gates track where the drone flies through two gates cyclically is done. (Figure \ref{fig:race track}) A first experiment aims to gather onboard data to be analyzed off-line. Hence, Opti-track system is used to provide accurate position measurements to make the loop closed. It should be noted that only in straight parts, the gate is in the drone's filed of view and the snake gate detection algorithm is done onboard, while the pose estimation and EKF are done off-board. The outer-loop controller is a PD controller combining Opti-track measurements to steer the drone to align with the center of the gate. In the arc parts, the gates are no longer available for navigation and the drone navigates itself to fly along an arc only by state prediction without the involvement of Opti-track, which will be explained in details in next section. The filtering result is shown in Figure \ref{fig:EKF result}. During the straight part (purple vision measurements), the EKF runs state prediction and measurement update loop and the estimated states curves (red) coincide with ground truth curves (blue) well. The error distributions between estimated states and ground-truth states are shown in Figure \ref{fig:EKF error}. All histograms are centered around 0 error. But there are still a few estimation errors above $0.2m$ in both x error and y error distribution which explains the fact that a few arcs end up at points which are more than 0.5m from target endpoint, which could be seen in next section. To make readers clearer to the experiment set up and result, a 3D ground truth and estimation result can be found in Figure{\ref{fig:3D trajectory}}   

\section{Control strategy}\label{sec:control_strategy}
Like classic control strategy of quadrotor, our control system is also divided into a inner-loop controller which stabilizes attitude of the quadrotor and a outer-loop controller which steers the quadrotor along the desired trajectory. For the inner-loop controller, an INDI controller is employed on-board\cite{Smeur2016}.

\begin{figure}[!h]
    \centering
    \includegraphics[scale=0.4]{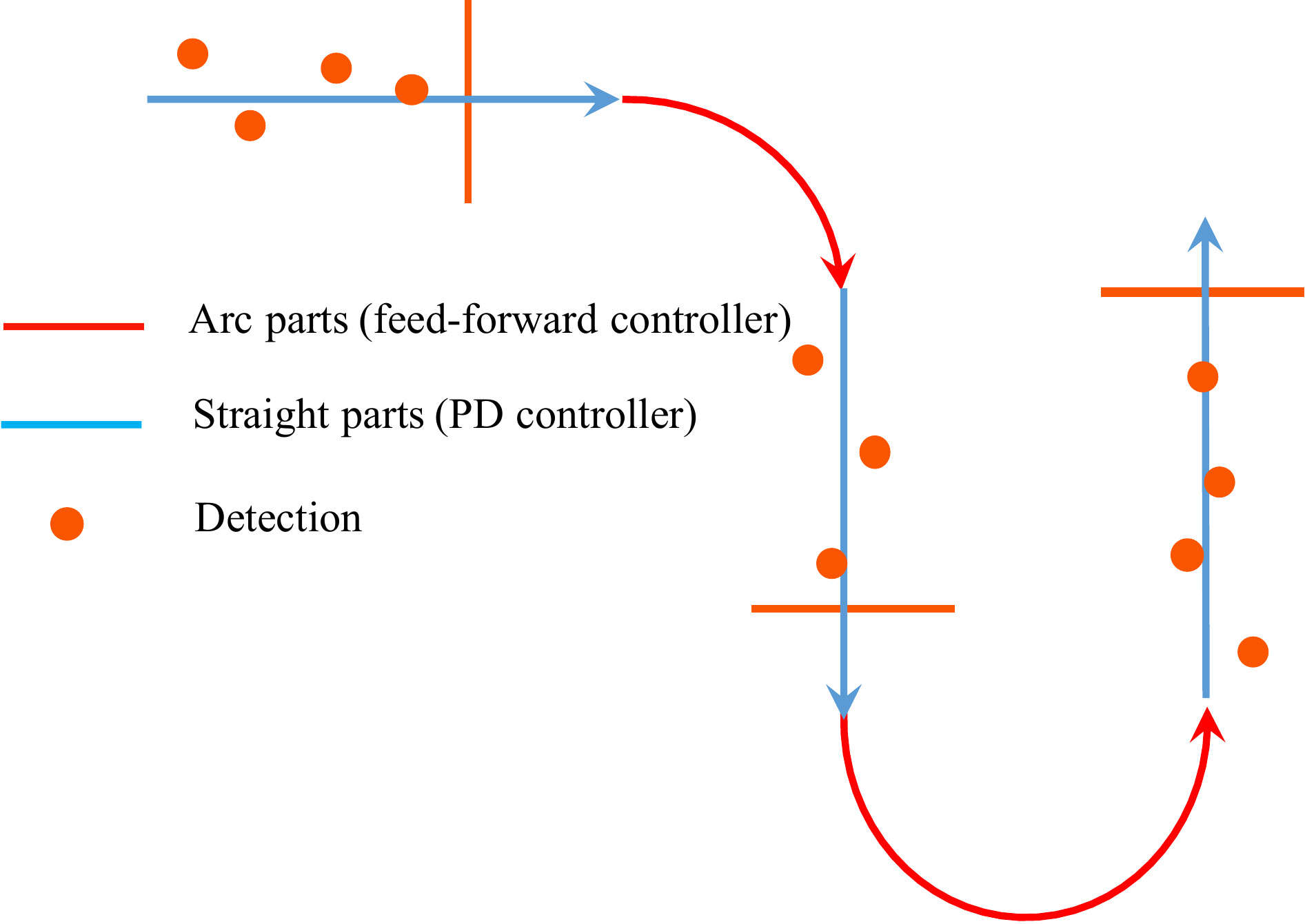}
    \caption{Two control strategies used in the experiment. When the drone faces the gates (straight parts), A PD controller combined with the Kalman filter is used to steer the drone to align with the gate. After passing through the gate, the drone switches to a feed-forward controller to fly an arc which ends in front of the next gate.}
    \label{fig:control strategy}
\end{figure}

For outer-loop control, we have two different control strategies for straight parts and arc parts respectively (Figure \ref{fig:control strategy}). During the straight part where the drone faces the gate and the gate is available for visual navigation, a PD controller is used to command a roll maneuver to steer the drone to align with the center of the gate while the pitch angle is fixed to a certain degree $\theta_0$ and the heading is fixed to the same direction as the gate.

\begin{align}
\begin{cases}
\phi_c = -k_p\hat{y}-k_d\dot{\hat{y}} \\
\theta_c =  \theta_0\\
\psi_c = 0^{\circ}
\end{cases}
\label{equ:feed_back control law}
\end{align}

where subscript $c$ means command and position $y$ is defined in local frame whose origin is fixed at the center of the gate.

\begin{figure}[!h]
    \centering
    \includegraphics[scale = 0.5,trim= 0cm 7cm 0cm 5cm,clip]{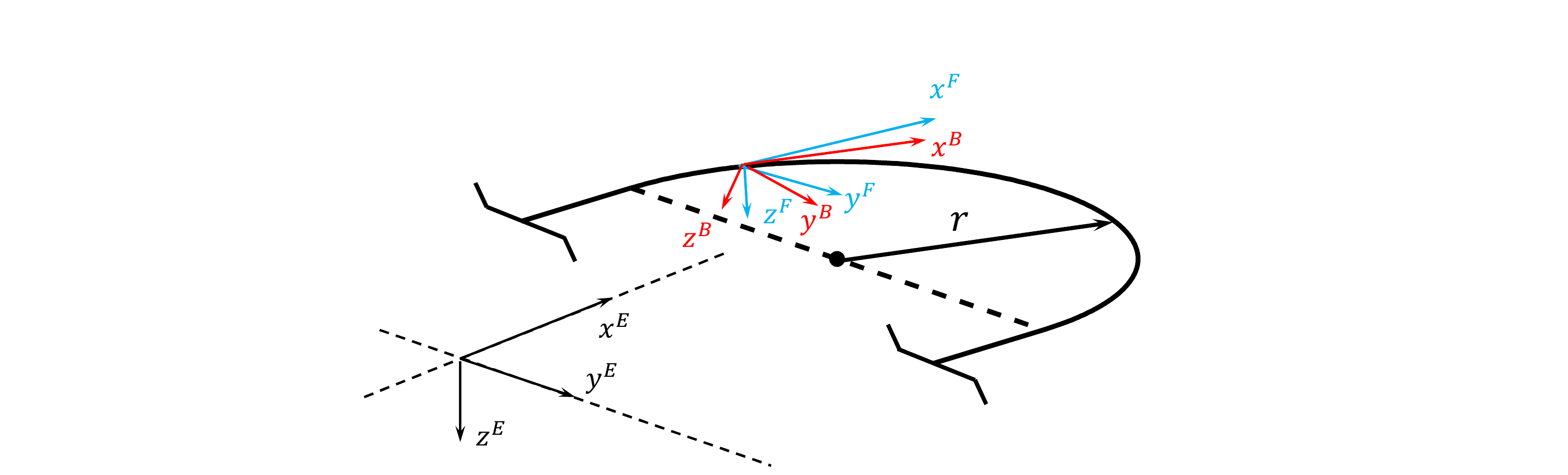}
    \caption{Body fixed earth frame $F$ whose origin $O^F$ is at the mass point of the drone, $X^F$ is along the heading of the quadrotor, $Z^F$ points to the earth. The rotation matrix from $E$ to $F$ is $\Re_E^F(\psi)$. The rotation matrix from $B$ to $F$ is $\Re_B^F(\phi,\theta)$ } 
    \label{fig:Frame F}
\end{figure}

At the point the drone flies through the gate, no position measurement is available. Thus, the outer-loop controller has to be switched to a pure feed-forward controller relying on state prediction to turn a coordinated arc which ends in front of the next gate. To derive the control law in the arc, we first introduce body fixed earth frame $F$ (Figure \ref{fig:Frame F}) whose origin $O^F$ is at the mass point of the drone, $X^F$ is along the heading of the drone, $Z^F$ points to the earth. In other words, the only non-zero Euler angle from $E$ to $F$ is yaw which is the same with the drone's yaw angle. To express Newton second law in $F$ we have

\begin{align}
\frac{\partial\mathbf{v}}{\partial t}\bigg|_{F}+\mathbf{\Omega}\times\mathbf{v}=\mathbf{F}
\label{equ:Newton in vector}
\end{align}

where $\frac{\partial\mathbf{v}}{\partial t}\bigg|_{F}$ is the derivative of $\mathbf{v}$ in $F$, $\mathbf{F}$ is the force acting on the drone and $\mathbf{\Omega}$ is angular velocity of Frame $F$ with respect to earth frame $E$. During the arc, the drone's heading is supposed to be tangent to the arc to maintain a zero sideslip turn, the angular velocity of $F$ with respect to $E$ should be 

\begin{align}
\mathbf{\Omega} = \begin{bmatrix}
0 \\ 0 \\ \dot{\psi}
\end{bmatrix} = \begin{bmatrix}
0 \\ 0 \\ \frac{v_x^F}{r}
\end{bmatrix}
\end{align}

To express equation \ref{equ:Newton in vector} in scalar form, we have

\begin{align}
\begin{bmatrix}
\frac{\partial v_x^F}{\partial t} \\ \frac{\partial v_y^F}{\partial t} \\ \frac{\partial v_z^F}{\partial t}
\end{bmatrix} = 
\Re_E^F\begin{bmatrix}
0 \\ 0 \\ g
\end{bmatrix} + 
\Re_B^F\begin{bmatrix}
0 \\ 0 \\ T
\end{bmatrix}+
\begin{bmatrix}
a_x^F \\ a_y^F \\ a_z^F
\end{bmatrix}-\begin{bmatrix}
0 \\ 0 \\ \frac{v_x^F}{r}
\end{bmatrix}\times
\begin{bmatrix}
v_x^F \\ v_y^F \\ v_z^F 
\end{bmatrix}
\label{equ:model in F}
\end{align}

where $T$ is the thrust of the drone and 

\begin{align}
\begin{bmatrix}
a_x^F \\ a_y^F \\ a_z^F
\end{bmatrix}=\Re_B^F
\begin{bmatrix}
k_x & 0 & 0 \\ 0 & k_y & 0 \\ 0 & 0 & k_z
\end{bmatrix}\Re_F^B
\begin{bmatrix}
v_x^F \\ v_y^F \\ v_z^F
\end{bmatrix}
\end{align}

During the arc, we would like to keep the altitude not changed, which in this frame means at the same height as at the start of the arc. Thus to make $\frac{\partial v_z^F}{\partial t}=0$ in equation \ref{equ:model in F}, we can have
\begin{align}
T=\frac{-g-a_z^F}{\cos\theta\cos\phi}
\label{equ:thrust in arc}
\end{align}

In  the arc, $\frac{\partial v_y^F}{\partial t}$ should be enforced to 0, substitute equation \ref{equ:thrust in arc} to the second line of equation \ref{equ:model in F}, we have,

\begin{align}
\phi_c = \tan^{-1}\frac{(a_y^F-\frac{{v_x^F}^2}{r})\cos\theta}{-g-a_z^F}
\end{align}

Similar to the straight part, pitch command $\theta_c$ is also fixed to a certain value. To conclude, during the arc maneuver, the control inputs are

\begin{align}
\begin{cases}
\psi_c(t) = \int_0^t\frac{{v_x^F(t)}}{r}dt \\
\phi_c(t) = \tan^{-1}\frac{(a_y^F(t)-\frac{{v_x^F(t)}^2}{r})\cos\theta_c(t)}{-g-a_z^F(t)} \\
\theta_c(t) = \theta_0
\end{cases}
\label{equ:control law}
\end{align}

\begin{figure}[!h]
    \centering
    \includegraphics[width=\textwidth]{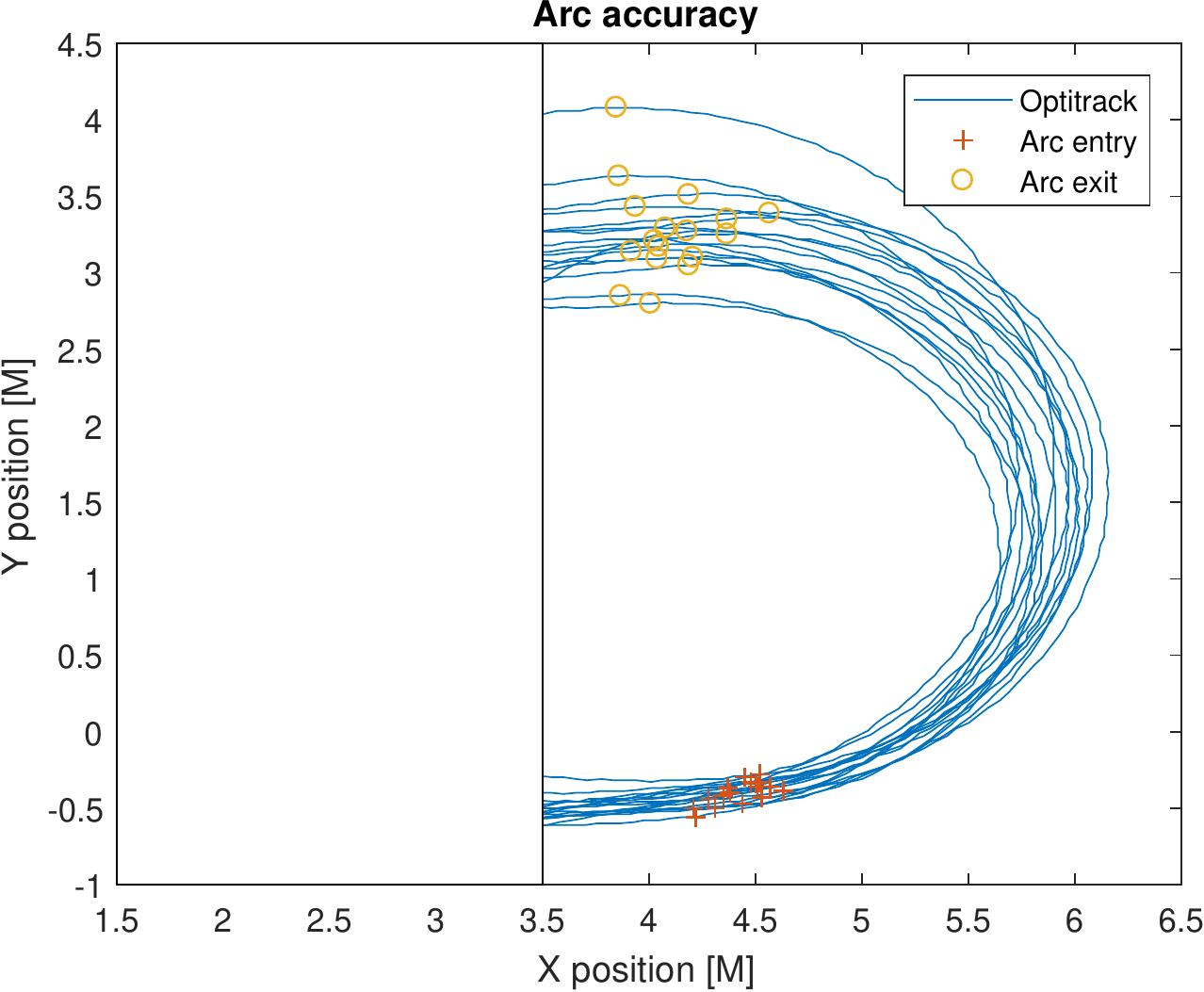}
    \caption{The flight test result of feed-forward control.The start points of the arcs (red points) slightly differ from each other because of the filter error. The end points of the arcs have a larger variance because the arc maneuvers are based on state prediction which is affected by model accuracy and initial state estimation.} 
    \label{fig:feed forward control}
\end{figure}

The flight test result can be found in Figure \ref{fig:feed forward control}. The drone enters the arc at red points and starts feed-forward control with the control strategy in equation \ref{equ:control law}. In a feed-forward arc maneuver, $\theta_c=-5^{\circ}$, $r=1.5m$ and each arc takes around $2s$. Before entering the arc, the drone is steered by the feedback control strategy in equation \ref{equ:feed_back control law}. At the same time, visual navigation is running to estimate the states of the drone which also tells the drone where to start to turn an arc. Thus in each lap, red points are slightly different from each other which is caused by filtering error. It could also be seen that the endpoints (yellow points) of arc maneuver has a distribution with larger variance compared to that at entry points. It is mainly because that state prediction in principle is an integration based method, which may be highly affected by the accuracy of initial states. In table \ref{tab:feed-forward control accuracy}, it is clear that the error at entry point in the x direction is much less than the one in the y direction. As a result, the error in the y axis at the endpoints is larger than that in the x axis. This error can also be caused by model inaccuracy and the disturbance during the arcs. Thus, the pure feed-forward control strategy is only effective for short time durations. In our case, $2s$ is enough to steer the drone to the next gate where visual navigation is available and feedback control strategy can be switched on again.      

\begin{table}[h]
\caption{Feed-forward control accuracy distribution} % title of Table
\centering % used for centering table
\begin{tabular}{c c c} % centered columns (4 columns)
\hline\hline %inserts double horizontal lines
 Axis & Entry speed variance $\sigma_v$ & Position error variance $\sigma_x$   \\ 
 X & $0.0043m/s$ & $0.0296m$ \\ 
 Y & $0.0106m/s$  & $0.8087m$\\
 \hline
\end{tabular}
\label{tab:feed-forward control accuracy} % is used to refer this table in the text
\end{table}

\begin{figure}[!h]
    \centering
    \includegraphics[width=0.7\textwidth]{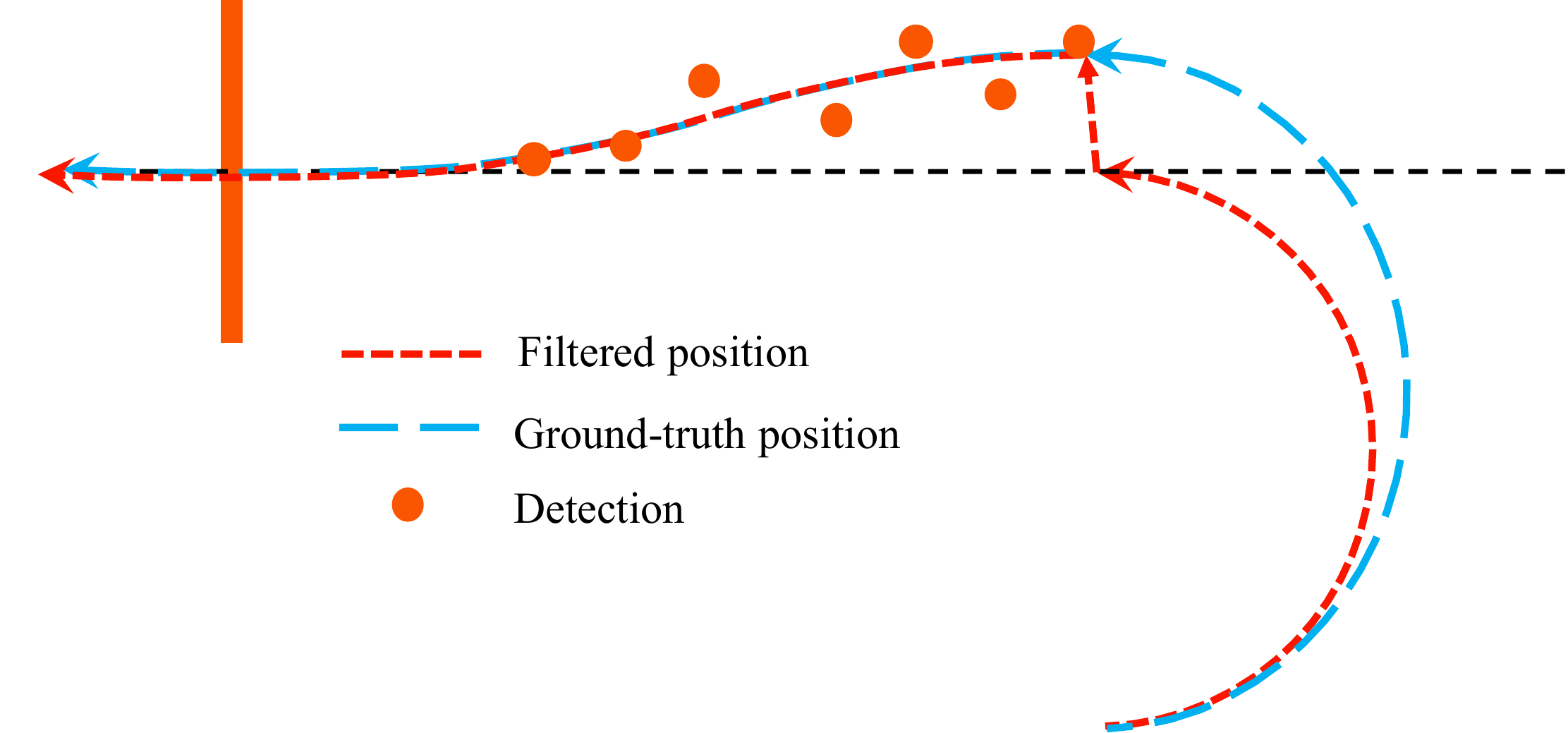}
    \caption{During the arc maneuver, the drone will not detect the gate. Thus, the state estimation is purely based on the prediction (red arc). However, due to the model inaccuracy and the sensors' bias, the predicted trajectory will diverge from the ground-truth trajectory (blue curve). After the turn, the drone will detect the gate again and the estimated position will jump to the ground-truth position. In fact, although there is a jump in the state estimation (red curve), the real-world trajectory should be continued (blue curve).}
    \label{fig:jump after arc}
\end{figure}

After the arc, the drone will detect the gate again and the detection will correct the filtering error. Thus, there will be a jump in the filtering result (Figure \ref{fig:jump after arc}). For the feed-back controller, the control target is to steer the drone to $y=0$. In fact, this is a simple step signal tracking or a way-point tracking problem. Simulations are done to check the feasibility of the proposed controller to steer the drone through the gate. The simplified drone model is 

\begin{figure}[h!]
\centering
\subfloat[$v_x=1.5m/s$]
{
  \includegraphics[scale = 0.3]{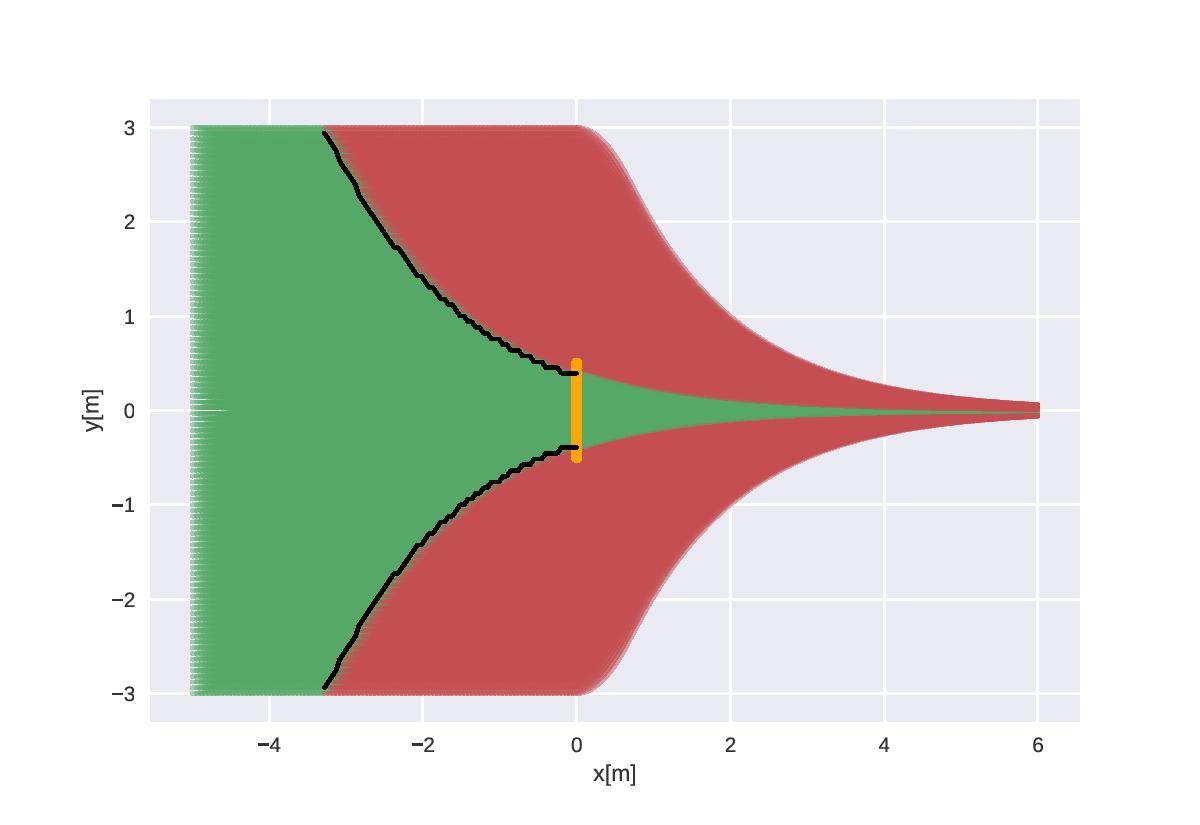}
}
\subfloat[$v_x=2m/s$]
{
  \includegraphics[scale=.3]{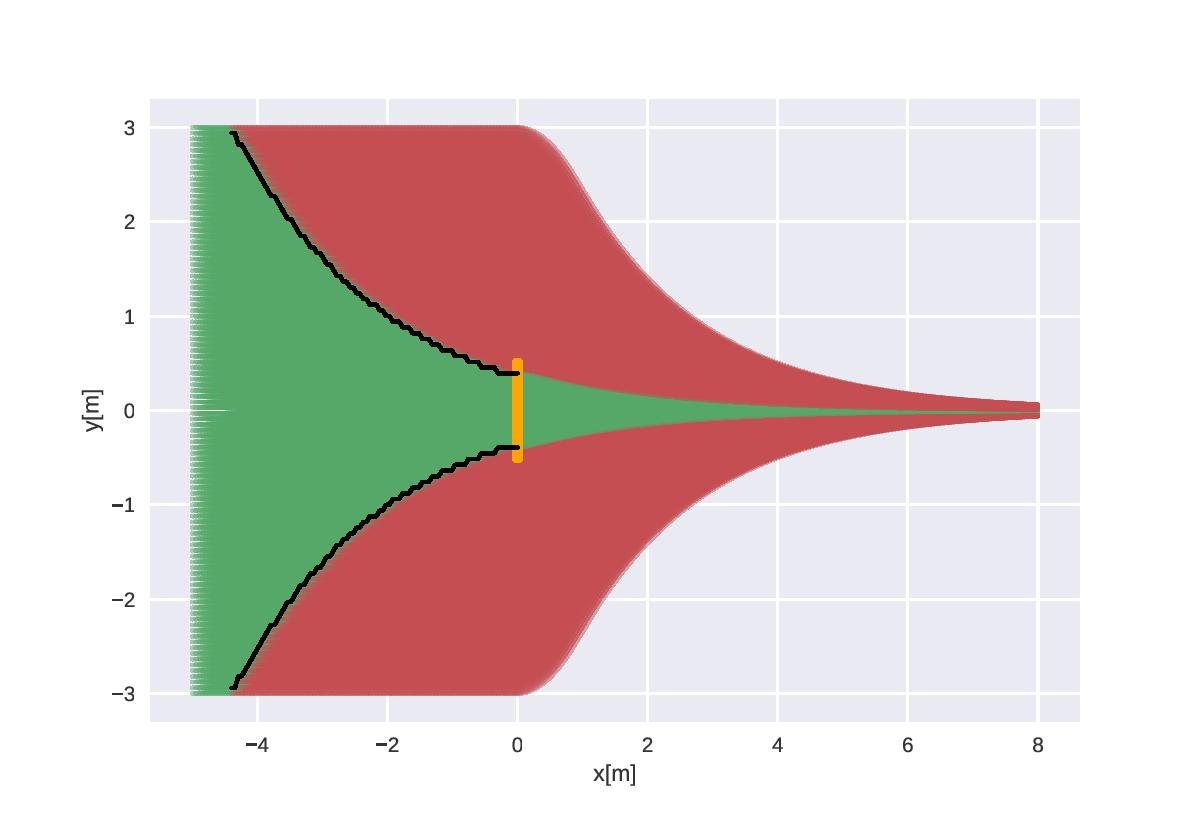}
}
\caption{The simulation result of the drone's passing through the gate. Whether the drone can pass through the gate depends on its initial position $x_0,y_0$ and its forward speed $v_x$. In each figure, $10000$ simulations are done with different initial points $x_0\in[-5m,0m],y_0\in[-3m,3m]$. The area to the left of the black curve is the set of the points, from which the drone can pass through the gate. Obviously, when the forward speed gets larger, the feasible initial points become less.}
\label{fig:sim_different_intial_point}
\end{figure}

\begin{align}
    \begin{cases}
    \dot{x} = v_x \\ \dot{y} = v_y \\ \dot{v}_y=g\tan{\phi}+k_yv_y\cos^2\phi
    \end{cases}
\end{align}
where $x$ and $y$ are the position of the drone and $v_y$ is the velocity of the drone in $y$ direction. In this model, we neglect $z$ because in the real-world flight, the altitude is controlled by a separate controller which can keep the altitude to be a constant. $v_x$ is the input of the model because in our real-world experiment setup, $\theta$ is set to be a constant which leads to a constant velocity in $x$ direction. $\phi$ is another input of the model. A PD controller is employed to steer the drone to $y=0$ by $\phi=k_v(k_p(0-y)-v_y)$, where $k_p = 1$ and $k_v=2$. The simulation result can be found in Figure \ref{fig:sim_different_intial_point}

% \begin{figure}[h!]
% \centering
% \subfloat[$v_x=1.5m/s$]
% {
%   \includegraphics[scale = 0.3]{control/control_convergence_1_5.pdf}
% }
% \subfloat[$v_x=2m/s$]
% {
%   \includegraphics[scale=.3]{control/control_convergence_2.pdf}
% }
% \caption{The simulation result of the drone's passing through the gate. Whether the drone can pass through the gate depends on its initial position $x_0,y_0$ and its forward speed $v_x$. In each figure, $10000$ simulations are done with different initial points $x_0\in[-5m,0m],y_0\in[-3m,3m]$. The area to the left of the black curve is the set of the points, from which the drone can pass through the gate. Obviously, when the forward speed gets larger, the feasible initial points become less.}
% \label{fig:sim_different_intial_point}
% \end{figure}

Figure \ref{fig:sim_different_intial_point} is the simulation result with the forward speed $v_x = 1.5m/s$ and $v_x=2m/s$. In each figure, $10,000$ trajectories are simulated with their own initial points $x_0\in[-5m,0m],y_0\in[-3m,3m]$. The points to the left of the black curves are the initial points from which the drone can pass through the gates. It can be seen that when the drone's speed gets higher, the number of the feasible initial points gets smaller. In other words, the drone needs more distance to adjust its position to pass through the gate. In our real-world experiment set up, for example, the forward speed is around $1.5m/s$ and the position error in $y$ axis is $0.8m$ as shown in Table \ref{tab:feed-forward control accuracy}. The drone needs a margin of $2m$ in $x$ direction to steer itself through the gate safely.

\section{Full track experiment setup and result}\label{sec:experiment}
\begin{figure}[h!]
\centering
\begin{tabular}{c c}
\subfloat[The first gate of the track]
{
  \includegraphics[scale=.3]{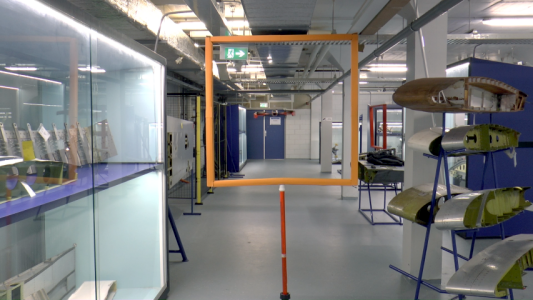}
}\hfill
&
\subfloat[The second gate of the track]
{
  \includegraphics[scale=.3]{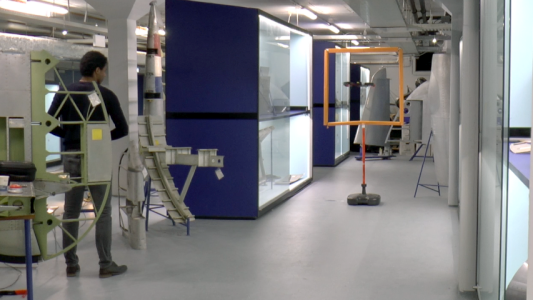}
}
\hfill \\
\subfloat[The third and fourth gate of the track]
{
  \includegraphics[scale=.3]{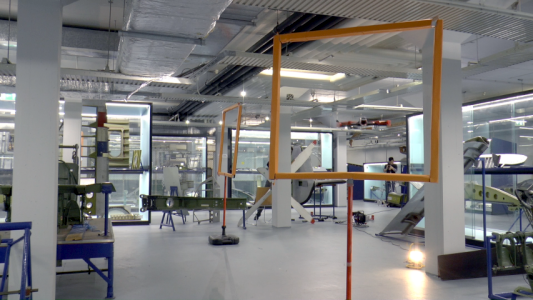}
} &
\subfloat[The fifth gate of the track]
{
  \includegraphics[scale=.3]{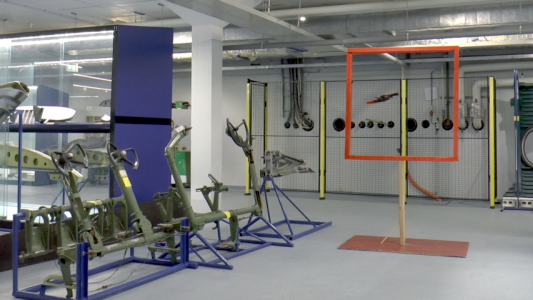}
} \hfill \\
\subfloat[Onboard snake gate detection]
{
  \includegraphics[scale=.25]{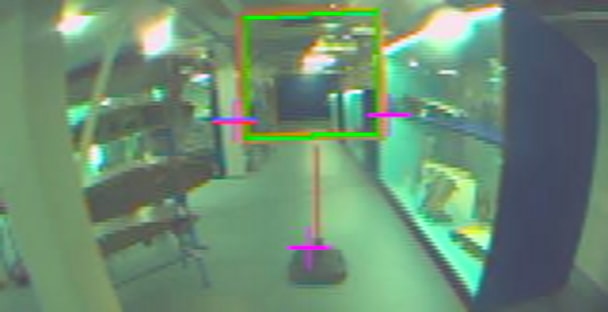}
}\hfill & 
\subfloat[Onboard histogram detection]
{
  \includegraphics[scale=.25]{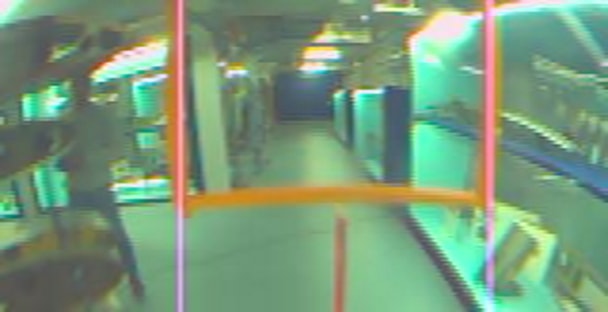}
}
\end{tabular}
\caption{Five gates are placed in a dense obstacle track. The gates are placed in narrow corridors and are surrounded by dense obstacles such as aircraft flaps, rudders and yokes. The first two row images are the environment around the gates and the last row are the onboard images with detection results.}
\label{fig:5 gates}
\end{figure}

In the previous sections, we have discussed the proposed visual navigation method and control strategies and the results of the experiments designed to verify our method in laboratory environment. In this section, we integrate all subsystems and move to a more challenging and realistic environment, a showroom in the basement of the Faculty of Aerospace Engineering, TU Delft where many aircraft components are displayed, to test the performance of our method. In this showroom, we placed five $1m\times1m$ gates in the corridor which is surrounded by dense showcases and aircraft components such as aircraft flaps, rudders, yokes and so on. The five gates are shown in Figure \ref{fig:5 gates}. Compared to the IROS 2017 autonomous drone racing, this track has smaller gates, much denser obstacles and the background of the gates is complex which in all put many challenges for the drone to fly the whole track fully autonomously.

\begin{figure}[!h]
    \centering
    \includegraphics[width=\textwidth]{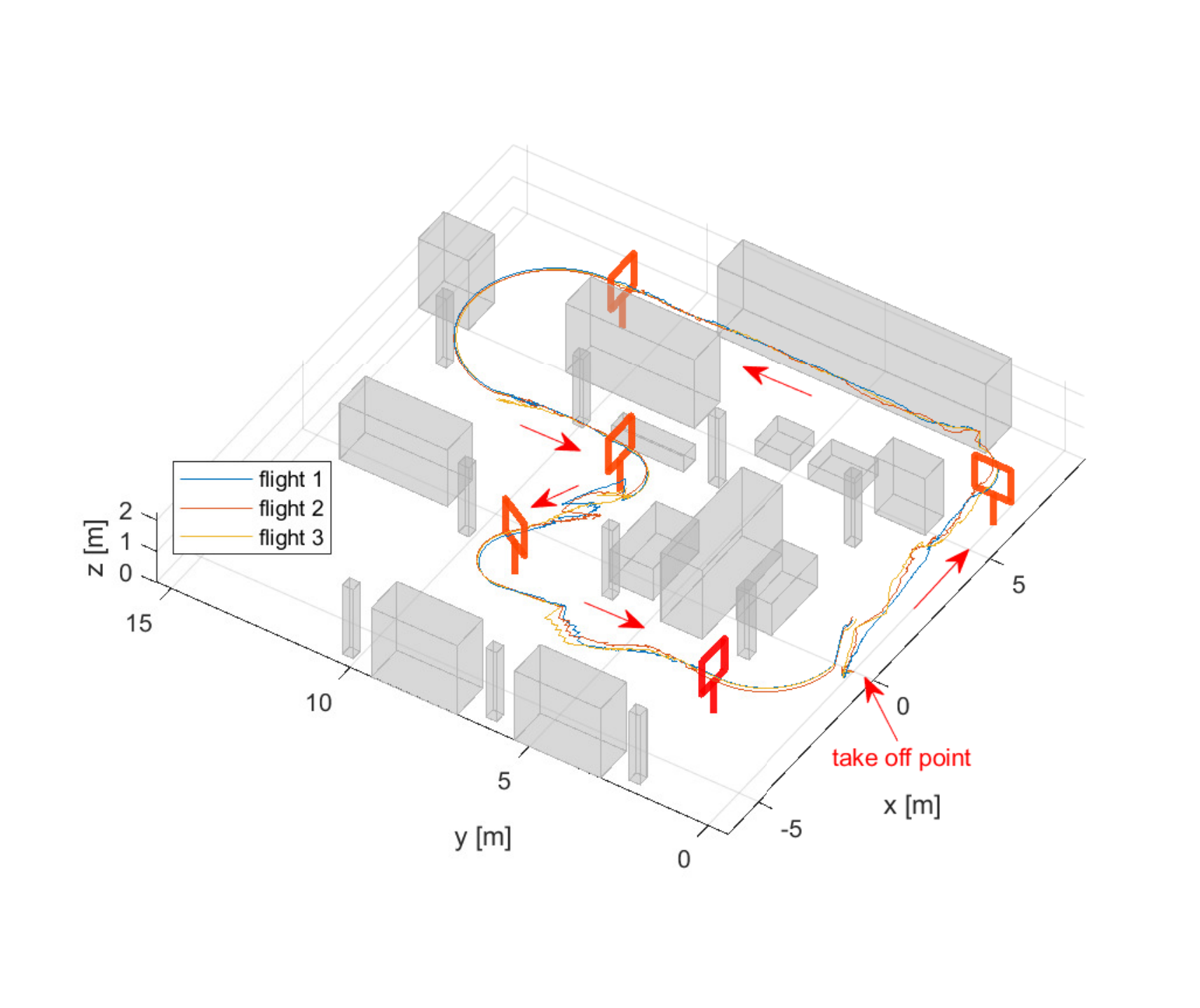}
    \caption{3 independent flight trajectories in the basement. It should be noted that these trajectories are the position estimation result of the flight instead of ground-truth trajectories.} 
    \label{fig:basement 3D}
\end{figure}

In this track, the drone takes off from ground and flies through the whole track with $\theta=-5^{\circ}$ or $\theta=-7^{\circ}$, which lead to the forward speed to be around $1.5m/s$ and $1.8m/s$ respectively, which is faster than the winner in autonomous drone race in 2016 who flew through 10 gates with $86s$ \cite{jung2018direct}, whose velocity is around $0.5m/s$. The onboard images and the flight result can be found in Figure \ref{fig:5 gates} and Figure \ref{fig:basement 3D}.

The environment is not equipped with a ground truth position system, therefore only estimated data is available. However, analyzing the estimated trajectory does give an insight of the flight and estimation performance in general. It can be observed that during some parts of the track some rapid changes in position occur. These jumps in position estimate occur once the next gate is first detected after a long period without seeing a gate. During this period the position estimation only relies on the integration of the drag based velocity. Errors in this prediction introduce an accumulating drift in the position estimate, which is corrected when a gate detection is available again. After the correction, the lateral position controller has enough time to steer the drone through the gate.

During the experiments, although in most cases the drone can pass through the gate, there are still some failure cases (the drone crashes to the gate). They are caused by non-detection of the gates or very late detection when the drone is already very close to the gate. In these two scenarios, the drone has to control itself purely based on prediction or the drone has no time to adjust its position. In our basement experiment, the poor quality of the onboard images leads to these non-detection problems. In terms of the open-loop control strategy, with the estimated linear aerodynamic model, we find that the control performance is very accurate in a short time. For example, after the second gate, there is a pole that is close to the arc (Figure \ref{fig:basement 3D}), but the drone never crashed into this pole.     

\section{Conclusion and future work}\label{sec:conclusion}
In this paper, we present a systematic scheme to accomplish the task of autonomous drone racing, as held by IROS in 2017. In our work, a novel and computationally efficient gate detection method is implemented onboard a Parrot Bebop 1 drone with all algorithms executed at 20 HZ frequency. With the detected gates, we employ a pose estimation scheme combining onboard AHRS estimation, which has higher accuracy than the commonly used P3P method. Then a more efficient Kalman filter is implemented onboard which converges faster than a traditional 15-states Kalman filter. In terms of the control strategy, a prediction-based feed-forward control strategy is used to control the drone to fly in the short time intervals without position measurements. And finally, the whole system is tested in a showroom with dense showcases and aircraft components. In this flight test, the average speed reached $1.5m/s$ which is higher than the speeds exhibited at the autonomous drone races in 2016 and 2017.

There are multiple directions for future work. For instance, the visual process is essentially based on color detection. Higher robustness in the visual processing may be reached by employing machine learning methods in computer vision. Also, a PD-controller is used to steer the drone through the gate, which makes the trajectory sub-optimal and can on the long term lead to overshoot. This can be improved, e.g., by utilizing optimal control methods. We hope that such future improvements will allow further augmenting the flight speed, hopefully approaching human pilot performance.

\section*{References}

\bibliography{elsarticle-template}

\section*{Appendex: Extended Kalman filter}
(1) Predict states based on equation \ref{equ:EKF model}

\begin{align}
\hat{\mathbf{x}}_{k|k-1}=\hat{\mathbf{x}}_{k-1}+\mathbf{f}(\hat{\mathbf{x}}_{k-1},\mathbf{u}_{k-1}){\rm T}
\end{align}

(2) Linearize and discretize the system

\begin{align}
\begin{split}
&\mathbf{F}_{k-1}=\frac{\partial}{\partial \mathbf{x}}\mathbf{f}(\mathbf{x}(t),\mathbf{u}(t))|_{{\mathbf{x}(t)=\hat{\mathbf{x}}}_{k-1}} \\
&\Phi_{k|k-1}\approx\mathbf{I}+\mathbf{F}_{k-1}{\rm T} \\
&\mathbf{H}_k=\frac{\partial}{\partial \mathbf{x}}\mathbf{h}(\mathbf{x}(t))|_{{\mathbf{x}(t)=\hat{\mathbf{x}}}_{k|k-1}} \\
%&\mathbf{\Gamma}_{k|k-1} \approx \mathbf{I}+\mathbf{H}_k{\rm T} \\
\end{split}
\end{align}

(3) Calculate prediction covariance matrix $\mathbf{P}_{k|k-1}$

\begin{align}
\begin{split}
\mathbf{P}_{k|k-1}= & \mathbf{\Phi}_{k|k-1}\mathbf{P}_{k-1}\mathbf{\Phi}_{k|k-1}^{\rm T}+\mathbf{Q}_{k-1}
\end{split}
\end{align}
where $\mathbf{Q}_{k-1}$ is system noise covariance matrix.

(4) Calculate Kalman gain and update prediction.

\begin{align}
\begin{split}
\delta\hat{\mathbf{x}}_k &= \mathbf{K}_k\left \{ \mathbf{Z}_k-\mathbf{h}[\hat{\mathbf{x}}_{k|k-1},k]\right \} \\
\mathbf{K}_k &=\mathbf{P}_{k|k-1}\mathbf{H}_k^{\rm T}[\mathbf{H}_k\mathbf{P}_{k|k-1}\mathbf{H}_k^{\rm T}+\mathbf{R}_k]^{-1} \\
\hat{\mathbf{x}}_k &= \hat{\mathbf{x}}_{k|k-1}+\delta\hat{\mathbf{x}}_k
\end{split}
\end{align}

where $\mathbf{R}_k$ is sensor noise covariance matrix.

\medskip

(5) Update the covariance matrix of state estimation error

\begin{align}
\mathbf{P}_k=(\mathbf{I}-\mathbf{K}_k\mathbf{H}_k)\mathbf{P}_{k/k-1}(\mathbf{I}-\mathbf{K}_k\mathbf{H}_k)^{\rm T}+\mathbf{K}_k\mathbf{R}_k\mathbf{K}_k^{\rm T}
\end{align}

\end{document}